\documentclass[final,1p,times]{elsarticle}

\usepackage{amssymb}

\usepackage{amsmath}
\usepackage{array}
\usepackage{pifont}
\usepackage{subcaption}
\usepackage{multirow}
\usepackage{etoolbox}
\usepackage{natbib}
\usepackage{color}
\usepackage[dvipsnames]{xcolor}
\usepackage{ulem}
\apptocmd{\sloppy}{\hbadness 10000\relax}{}{}
\usepackage{enumitem}
\usepackage{algorithm}
\usepackage{algpseudocode}
\usepackage{textcomp}
\usepackage{amsmath}
\usepackage{hyperref}
\usepackage{hyperref}

\newcommand{\cmark}{\color{ForestGreen}{\ding{51}}}%
\newcommand{\xmark}{\textcolor{red}{\ding{55}}}%

\newcommand{\fetwo}{FE$^\mathrm{2}$}
\newcommand{\ie}{\textit{i}.\textit{e}. }

\begin{document}

\begin{frontmatter}



\title{A Microstructure-based Graph Neural Network for Accelerating Multiscale Simulations}

%

\author{J. Storm}
\author{I. B. C. M. Rocha}
\author{F. P. van der Meer}

\address{{Delft University of Technology, Department of Civil Engineering and Geosciences}, {P.O. Box 5048}, 2600GA, Delft, {The Netherlands}}


\begin{abstract}
Simulating the mechanical response of advanced materials can be done more accurately using concurrent multiscale models than with single-scale simulations.
However, the computational costs stand in the way of the practical application of this approach.
The costs originate from microscale Finite Element (FE) models that must be solved at every macroscopic integration point.
A plethora of surrogate modeling strategies attempt to alleviate this cost by learning to predict macroscopic stresses from macroscopic strains, completely replacing the microscale models.
In this work, we introduce an alternative surrogate modeling strategy that allows for keeping the multiscale nature of the problem, allowing it to be used interchangeably with an FE solver for any time step.
Our surrogate provides all microscopic quantities, which are then homogenized to obtain macroscopic quantities of interest.
We achieve this for an elasto-plastic material by predicting full-field microscopic strains using a graph neural network (GNN) while retaining the microscopic constitutive material model to obtain the stresses.
This hybrid data-physics graph-based approach avoids the high dimensionality originating from predicting full-field responses while allowing non-locality to arise.
In addition, this approach introduces beneficial inductive bias to the model by encoding microscopic geometrical features.
By training the GNN on a variety of meshes, it learns to generalize to unseen meshes, allowing a single model to be used for a range of microstructures.
The embedded microscopic constitutive model in the GNN implicitly tracks history-dependent variables and leads to improved accuracy.
While the microscopic stresses are fully dependent on the microscopic strains, we found it crucial to include both microscopic strains and stresses in the loss function.
We demonstrate for several challenging scenarios that the surrogate can predict complex macroscopic stress-strain paths.
As the computation time of our method scales favorably with the number of elements in the microstructure compared to the FE method, our method can significantly accelerate \fetwo{} simulations.
\end{abstract}




\end{frontmatter}


\section{Introduction}\label{sec1}

There is a global demand for reducing the material cost of structures.
One method to achieve this is by creating structures with highly-tailored materials at various scales.
However, we currently lack the ability to accurately model the mechanical response of these materials.
We typically model these applications using the finite element (FE) method, a technique for solving partial differential equations.
The type of material used influences the constitutive relation between stresses and strains.
For materials with properties described at a much smaller scale (microscale) than the physical system (macroscale), such as fiber-reinforced polymers, often no accurate analytical relation is known.
Instead, concurrent multiscale techniques, such as \fetwo, can be used.
In \fetwo, the constitutive relations are described at the microscale where they are better known, and a full microscopic model is coupled to each integration point of the macroscopic model.

However, multiscale modeling has a high computational cost, therefore data-driven surrogates have been proposed to replace the microscale model directly.
Examples of these surrogates are Recurrent Neural Networks (RNNs)~\citep{ghavamian2019accelerating, borkowski2022recurrent, wu2020recurrent}, temporal convolutional networks (TCNs)~\citep{wang2022deep, abueidda2021deep} and attention-based architectures~\citep{wang2020general}.
These purely data-driven models require a substantial amount of data from expensive microscale simulations to generalize to unseen situations, if they can generalize at all.
Introducing physics back into the model can partly overcome this limitation, examples being learning to predict material invariants~\citep{linka2021constitutive}, using the invariants themselves as inputs~\citep{klein2022finite}, or predicting the internal variables of a constitutive relation~\citep{masi2021thermodynamics}.
Including the microscopic material model directly into the data-driven surrogate reduces the necessary training data and enables phenomena such as unloading to be captured without training for it~\citep{maia2022physically, liu2019deep, rocha2023machine}.

To obtain training data for these methods, microstructures are simulated for a wide variety of load scenarios.
There are different sampling approaches for selecting load paths, such as those from actual \fetwo{} simulations~\citep{ghavamian2019accelerating}, monotonically increasing paths~\citep{abueidda2021deep}, and Gaussian process (GP)-based paths~\citep{abueidda2021deep, maia2022physically}.
Each simulated microstructure contains a rich stress field of data, which is homogenized to a single stress vector to train the models.
This means that when using these macroscale-based surrogates, it is impossible to obtain full-field microscopic quantities.
Having full-field information could for example be used in an active-learning setting by switching to an FE solver for a few time steps.
However, incorporating full-field information in the surrogate model is not trivial with traditional data-driven models.

Next to developments of surrogates predicting a homogenized response for multiscale simulations, numerous works on data-driven techniques focus on predicting full-field single-scale FE problems.
Two related challenges here are making a model capable of handling arbitrary domains and avoiding the curse of dimensionality.
Physics-informed neural networks (PINNs) embed the partial differential equation (PDE) equilibrium in the loss function, learning based on a few examples and generalizing well~\citep{raissi2019physics, karniadakis2021physics}.
PINNs and graphs can be combined to get a method closely resembling the FE method, making it possible to minimize the residuals of the weak form of a PDE for linear elasticity~\citep{PIGalerkingGNN}.
Recently, the development of operators has gained a lot of attention.
Operators are analogous to learning an entire family of PDEs rather than a single instance of the equation.
DeepONet, the Fourier neural operator, and the graph neural operator are notable examples~\citep{lu2019deeponet, li2020fourier, li2020neural}.
These operators could have been suitable alternatives for this study.

Alternatively, convolutional neural networks (CNNs) and U-net are other data-driven models that have been used to predict quantities based on a full-field domain~\citep{cang2018improving, abueidda2019prediction} and generate a full-field solution for elastic cases~\citep{krokos2022bayesian, gupta2023accelerated, aldakheel2023efficient}.
Reduced order models have recently been combined with RNNs to recover the full-field solution in a multiscale simulation~\citep{vijayaraghavan2023data}.
While CNNs are limited to grid-structured data, graph neural networks (GNNs) can deal with arbitrary graphs, making them well-suited for mesh-based structures~\citep{krokos2022graph}.
Pfaff et al. propose MeshGraphNets, a GNN with an Encode-Process-Decode architecture followed by an integrator, and use it to predict the dynamics of physical systems such as aerodynamics and structural mechanics~\citep{pfaff2020learning}.
This GNN-based method predicts the relevant quantity for each element based on information of the surrounding elements.
Spreading this information around the mesh happens using Message Passing Layers (MPLs).
By increasing the number of MPLs, information spreads faster, but too many MPLs can lead to all representations becoming very similar, a problem known as over-smoothing~\citep{li2018deeper}.
Both dynamic and static time-independent problems can require fast propagation of information, and adding multiple resolutions of nodes or random augmentations allows information to spread with fewer MPLs, improving accuracy~\citep{li2020multipole, multiscaleGNN, cao2022bi, fortunato2022multiscale, gladstone2023gnn}.
These GNNs for predicting physical systems generally only predict a single time step ahead during training.
To prevent the predictions from diverging during inference over multiple time steps, a Gaussian or prediction-based noise can be added to the inputs during training~\citep{pfaff2020learning,brandstetter2022message}.
In addition, GNNs have been used to capture multiple steps of linear hardening plasticity under unidirectional tension for simple 1-fiber microstructures~\citep{maurizi2022predicting}, to create a lower-dimensional embedding of the plastic state~\citep{vlassis2023geometric}, and for crystal plasticity in 3D microstructures~\citep{frankel2022mesh}.
However, none of the aforementioned models is capable of obtaining full-field microscopic quantities for arbitrary inelastic load paths for microstructures with any distribution.

In this work, we propose a surrogate model that makes full-field microscopic predictions for multiscale simulations.
We create a dual graph over the mesh of the microstructure — essentially connecting the integration points — and opt for a GNN model.
The direct outputs are microscopic strains, and we embed the microscopic material constitutive relations inside the model to obtain stresses.
We then in turn use standard computational homogenization to obtain macroscopic stresses.
By including the material model, and training on both the strains and the stresses, we learn to predict complex load scenarios for arbitrary meshes more accurately than without the material model.
Instead of making one-step ahead predictions during training and adding noise to the inputs to improve stability, we train on predicting multiple time steps consecutively.
The surrogate acts similarly to an autoregressive model by using its strain predictions as inputs for the next time step, being analogous to an RNN with strong physical interpretability due to the material model implicitly updating and preserving history-dependent variables between time steps.
Furthermore, by introducing robust geometric inductive bias, our surrogate can successfully generalize to unseen mesh sizes.

We organize the paper as follows.
Section~\ref{sec:background} provides the essential background information on concurrent multiscale simulations and GNNs.
In Section~\ref{sec:GNN_surrogate}, we introduce our GNN-based surrogate model and provide a comprehensive description of its architecture and the process of training it.
Section~\ref{sec:monotonic} presents a test of the model on a scenario with monotonic loading conditions.
To do so, we start the section with an in-depth discussion on model selection, a critical aspect of developing a deep learning surrogate that is frequently left unaddressed.
Drawing parallels with the study by~\citep{maia2022physically}, we examine if the model can predict unloading without being specifically trained for it.
This motivates the subsequent Section~\ref{sec:nonmono}, where we train the model on non-monotonic data.
Then, we explore the impact of the embedded material model, demonstrate the model's capability to make predictions for various mesh sizes and for more time steps than the model has seen during training, and compare its computational cost to classical FE simulations.
Finally, Section~\ref{sec:conclusion} presents our conclusions.

\section{Background}\label{sec:background}
This section introduces single-scale FE methods before discussing multiscale FE methods and their challenges, which surrogate models aim to overcome.
Then, we present a brief overview of GNNs, which we use as the basis for the surrogate model.

\subsection{Single-scale FE analysis}
We aim to solve a PDE on a domain $\Omega$ subject to boundary conditions, finding the resulting displacement field $\mathbf{u}$.
The domain $\Omega$ is discretized into a number of elements, and the solution in this domain must satisfy:
\begin{equation}\label{eq:sig_constraint}
    \mathbf{\nabla} \cdot \boldsymbol{ \sigma}^{\Omega} = \mathbf{0},
\end{equation}
where $\boldsymbol{\sigma}$ is the stress and $\mathbf{\nabla} \cdot$ indicates the divergence operator.
We satisfy this relation using the following equations.
The displacement is related to the strain $\varepsilon$ as:
\begin{equation}\label{eq:eps_constraint}
    \boldsymbol{\varepsilon}^{\Omega} = \dfrac{1}{2} \left( \mathbf{\nabla u}^{\Omega} + \left( \mathbf{ \nabla u}^{\Omega} \right)^T \right),
\end{equation}
and the strain and stress relate via a constitutive model:
\begin{equation}\label{eq:material_alpha}
    \boldsymbol{\alpha}_{t}^{\Omega} = \mathbf{\mathcal{A}}^{\Omega} \left(  \boldsymbol{\varepsilon}^{\Omega}_t, \boldsymbol{\alpha}_{t-1}^{\Omega} \right)
\end{equation}
\begin{equation}\label{eq:material_sigma}
    \boldsymbol{\sigma}_{t}^{\Omega} = \mathbf{\mathcal{S}}^{\Omega} \left(  \boldsymbol{\varepsilon}^{\Omega}_t, \boldsymbol{\alpha}_{t}^{\Omega}  \right).
\end{equation}

For history-dependent problems, an internal material variable $\boldsymbol{\alpha}$ is used which evolves over time, for example, to consider elasto-plastic problems.
The functions $\mathbf{\mathcal{A}}^{\Omega}$ and $\mathbf{\mathcal{S}}^{\Omega}$ are material-dependent expressions.
In the nonlinear case, the resulting displacement $\mathbf{u}$ cannot be solved for directly, and an iterative Newton-Raphson optimization scheme is used instead to find a solution that satisfies Equations (\ref{eq:sig_constraint}) to (\ref{eq:material_sigma}).

\subsection{Concurrent multiscale analysis}\label{sec:multiscale}

\begin{figure}[htbp]
\centering
\includegraphics[width=.45\textwidth]{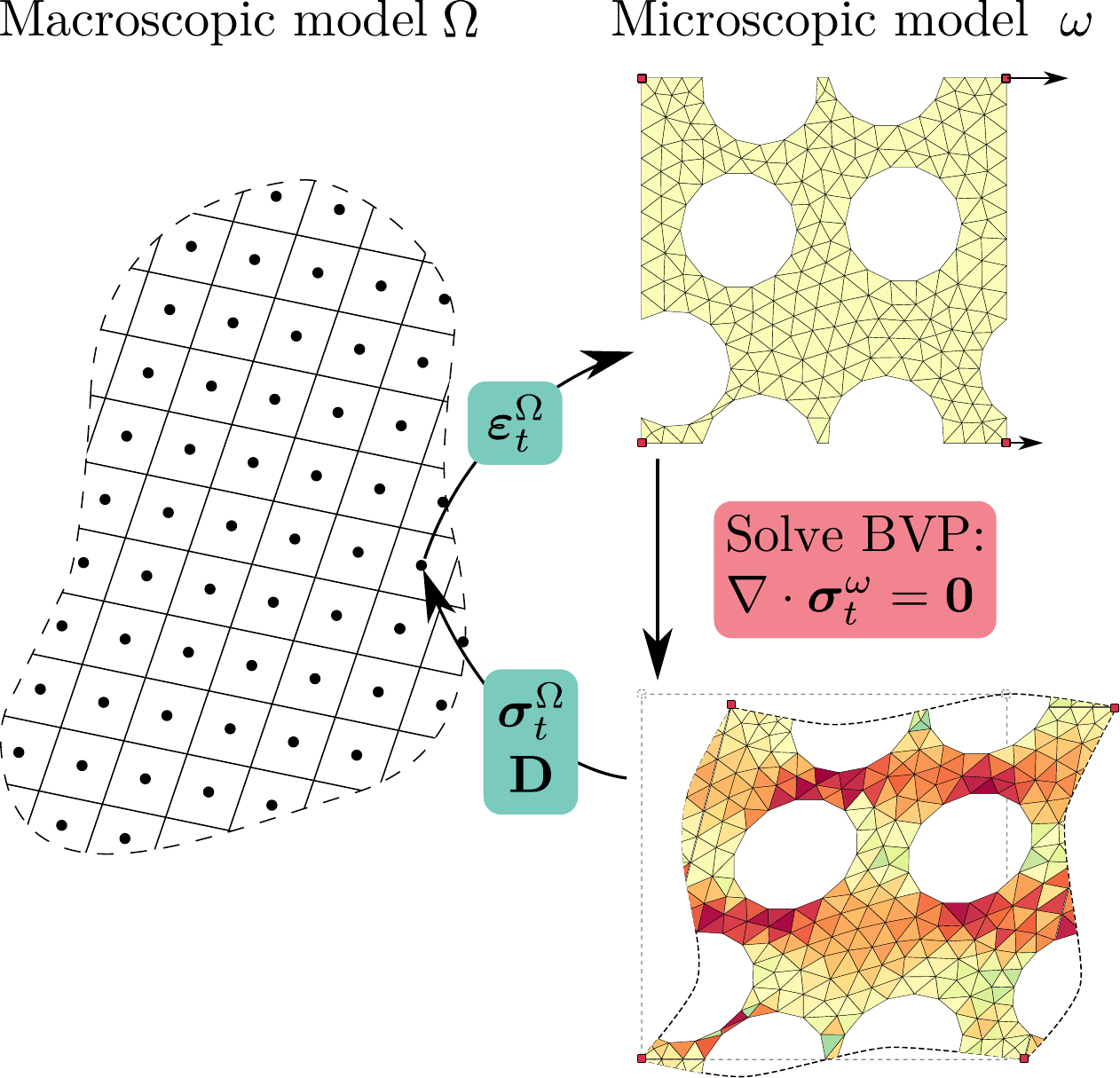}
\caption{An overview of an \fetwo{} simulation. A microscopic model $\omega$ is solved for every integration point of the macroscopic model $\Omega$. }
\label{fig:fe2overview}
\end{figure}

For composite materials with multiple constituents, a direct numerical simulation (DNS) approach involves explicitly modeling the microstructure and assigning different constitutive models for different points in $\Omega$.
However, if the microstructural length scales are much smaller than the size of the domain of interest, DNS becomes extremely computationally expensive.
In such cases, homogenization is necessary.
Finding accurate expressions for $\mathbf{\mathcal{A}}^{\Omega}$ and $\mathbf{\mathcal{S}}^{\Omega}$ can be challenging.
Alternatively, a concurrent multiscale analysis can be performed, where we replace the analytical expressions for $\mathbf{\mathcal{A}}^{\Omega}$ and $\mathbf{\mathcal{S}}^{\Omega}$ with solving a boundary value problem on a periodic microscopic domain $\omega$, as depicted in Figure \ref{fig:fe2overview}.
This approach is valid if the scales can be separated, \ie if $\omega \ll \Omega$.
To connect these scales, we assume that the microscopic displacements $\mathbf{u}^{\omega}$ can be related by a linear contribution proportional to the macroscopic strains $\boldsymbol{\varepsilon}^{\Omega}$ and a periodic fluctuation term $\tilde{\mathbf{u}}^{\omega}$ as:
\begin{equation}\label{eq:displacement}
    \mathbf{u}^{\omega} = \boldsymbol{\varepsilon}^{\Omega} \mathbf{x}^{\omega} + \tilde{\mathbf{u}}^{\omega},
\end{equation}
where $\mathbf{x}^{\omega}$ is the microscopic coordinate vector.
Eqs. (\ref{eq:sig_constraint}-\ref{eq:material_sigma}) then hold for the microscale $\omega$ where the relations for each constituent are accurately known.
Periodic boundary conditions are constructed by relating the nodal displacements on the boundary of $\omega$ with the nodes on the opposite side of the domain.
After the microscopic problem is solved, the stresses are homogenized to obtain the macroscopic quantity of interest:
\begin{equation}\label{eq:homog}
    \boldsymbol{\sigma}^{\Omega} = \dfrac{1}{\vert \omega \vert} \int_{\omega}^{}  \boldsymbol{\sigma}^{\omega} d\omega .
\end{equation}
The microscale problem together with the scale transition in Eqs. (\ref{eq:displacement}) and (\ref{eq:homog}) take the place of the constitutive model (\ref{eq:material_alpha}-\ref{eq:material_sigma}) in the macroscopic problem.

While the computational cost can be significantly reduced with respect to DNS, simulating these microscale models for all integration points for every macroscopic time step is still prohibitively expensive for most problems.
Therefore, attempts have been made to replace the microscale simulations with surrogate models, which are much faster to evaluate.
To train these surrogates, data from these microscale models is required.
Obtaining this data can be a costly process in and of itself, making it desirable to minimize the amount of data the surrogate requires.

\subsection{Graph Neural Networks}
The model architecture used in this work is based on a Graph Neural Network (GNN).
GNNs are used for various applications where graph-type data is available, and numerous different GNN-based architectures exist.
Here, we will provide a concise introduction to the message-passing GNN as is used in this work, for a more thorough introduction we refer to~\citep{hamilton2020graph}.

A graph $\mathcal{G} = (V, E)$ consists of vertices (or nodes) $V$ linked with edges $E$.
The graph can contain information on the edges ($\mathbf{e}_{ij}$), at the nodes ($\mathbf{v}_i$), and at the level of the graph itself, depending on the application.
The basic building block of a GNN is a message passing layer (MPL), allowing it to pass information from the nodes and edges around the graph.
First, based on all edge states $\mathbf{e}_{ij}$ and nodal states $\mathbf{v}_{i}$, the corresponding directional edge messages $\mathbf{m}_{ij}$ are computed:
\begin{equation}\label{eq:edge_update}
    \mathbf{m}_{ij} = F^E ( \mathbf{e}_{ij}, \mathbf{v}_i, \mathbf{v}_j )
\end{equation}
After updating the messages, every node aggregates the incoming messages from its connected edges (its neighborhood $\mathcal{N}$).
\begin{equation}\label{eq:aggregation}
    \mathbf{m}_{i} = F^{A} ( \mathbf{m}_{ij} \mid j \in \mathcal{N} )
\end{equation}
The aggregation function $F^{A}$ is a symmetric operation that can handle incoming messages from an arbitrary number of edges.
In practice, this often results in aggregation functions taking the mean or sum of incoming messages.
Based on the aggregated messages, the state of each node is updated:
\begin{equation}\label{eq:update}
    \mathbf{v}_{i} = F^{U} ( \mathbf{v}_i, \mathbf{m}_i ).
\end{equation}
The functions $F^E$, $F^{A}$, and $F^{U}$ are shared between all nodes and edges and generally operate on all node and edge tensors at once.
Eqs. (\ref{eq:edge_update}-\ref{eq:update}) represent one MPL, and multiple MPLs can be stacked to process information between more indirectly connected nodes.
The number of MPLs is thus a hyperparameter, and when using multiple MPLs, the functions $F^E$, $F^{A}$, and $F^{U}$ can either be either unique functions or shared between all layers.

\section{GNN-based surrogate model}\label{sec:GNN_surrogate}
In this work, we create a GNN-based surrogate model with the aim of accelerating \fetwo{} simulations.
By using a GNN, we can make full-field predictions while avoiding the high dimensionalities that arise from predicting a full-field response as in a standard NN and giving more flexibility than the grid-based approach of a CNN.
Furthermore, the mesh is used by the model as an inductive bias, allowing it to generalize to new meshes.
In this section, the details of how we construct this GNN with an embedded material model are presented.
We start by introducing the features of the graph, after which we describe the precise architecture and training procedure.

\subsection{Graph creation}
To use a GNN as a surrogate constitutive model for \fetwo{} simulations, we create a dual graph over the mesh, connecting all the integration points.
This is possible because we use elements with a single integration point per element.
From this point, we use the word "nodes" to refer to the vertices of the graph, not the nodes of the mesh, unless specified otherwise.
Based on this graph, a number of features are inserted for every node.
The current microscopic strain state $\boldsymbol{\varepsilon}^{\omega}_{t}$, the internal variables $\boldsymbol{\alpha}^{\omega}_{t}$, the macroscopic strains $\boldsymbol{\varepsilon}^{\Omega}_{t+1}$, and any additional geometric features $ \mathbf{g}$ are the inputs, which are encoded to $\mathbf{v}_i$.
The macroscopic strains $\boldsymbol{\varepsilon}^{\Omega}_{t+1}$ are the same for all nodes.
The edge features $\mathbf{e}_{ij}$ are the positional differences in $x$ and $y$ directions between the nodes it connects: $[x_i-x_j, y_i-y_j ]$.
With bidirectional edges, this thus results in $\mathbf{e}_{ji}=-\mathbf{e}_{ij}$.
We are able to retain periodicity by adding extra edges connecting the nodes at opposite sides of the microstructure.
The distances for these edges are computed with phantom nodes moved by the length of the microstructure in the corresponding direction, but otherwise, these edges are treated identically to other edges.

Although many GNN-based models for simulating physical processes impose boundary conditions based on forces and displacements on the mesh nodes, this is not straightforward to apply here.
The boundary condition directly influences the complete microstructure in the first step, requiring an infeasible number of MPLs to propagate this information to all elements.
We avoid moving to displacements and forces and instead directly input the macroscopic strains $\boldsymbol{\varepsilon}^{\Omega}_{t+1}$ uniformly to all nodes.
This allows us to only consider the microscopic integration points and work with stresses and strains, which is advantageous as embedding the material models is then straightforward.

\begin{figure}[htbp]
\centering
\includegraphics[width=0.35\textwidth]{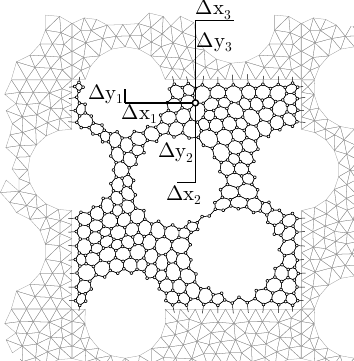}
\caption{A dual graph is created over the mesh, coinciding with the integration points.
Each node of this graph has unique features providing information about a number of the closest surrounding voids through their distances $\Delta x$ and $\Delta y$.
We consider the nine closest voids, but we only show distances to three here for clarity.
Periodicity is considered in both the void features and the edge connections - nodes at opposite sides of the microstructure are connected.}
\label{fig:graph_overview}
\end{figure}

As all other nodal features are the same for the initial time step, the locality in the response of the model is made possible by including geometric features.
These geometric features $\mathbf{G}$ are the $\Delta x$ and $\Delta y$ distance from each node to the center of a fixed number of voids, as visualized in Figure~\ref{fig:graph_overview}.
We are able to consider more voids as features than there are voids in the microscale mesh by accounting for periodicity.
In practice, we tile each microstructure in a grid and use a K-nearest neighbors algorithm for each node in the center mesh with respect to the voids.
This enables us to include small microstructures with only a few voids, which are cheap to obtain, while still having a sufficient number of features.

\subsection{Architecture}\label{sec:architecture}

\begin{figure}[tbp]
\centering
\includegraphics[width=0.45\textwidth]{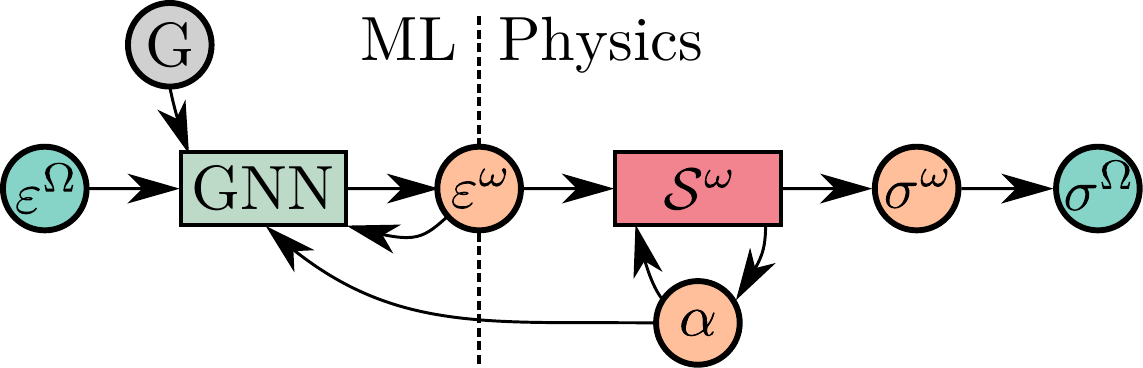}
\caption{Abstract overview of the multi-step GNN predictions.}
\label{fig:abstractGNN}
\end{figure}

\begin{figure}[htbp]
    \centering
    \includegraphics[width=.9\textwidth]{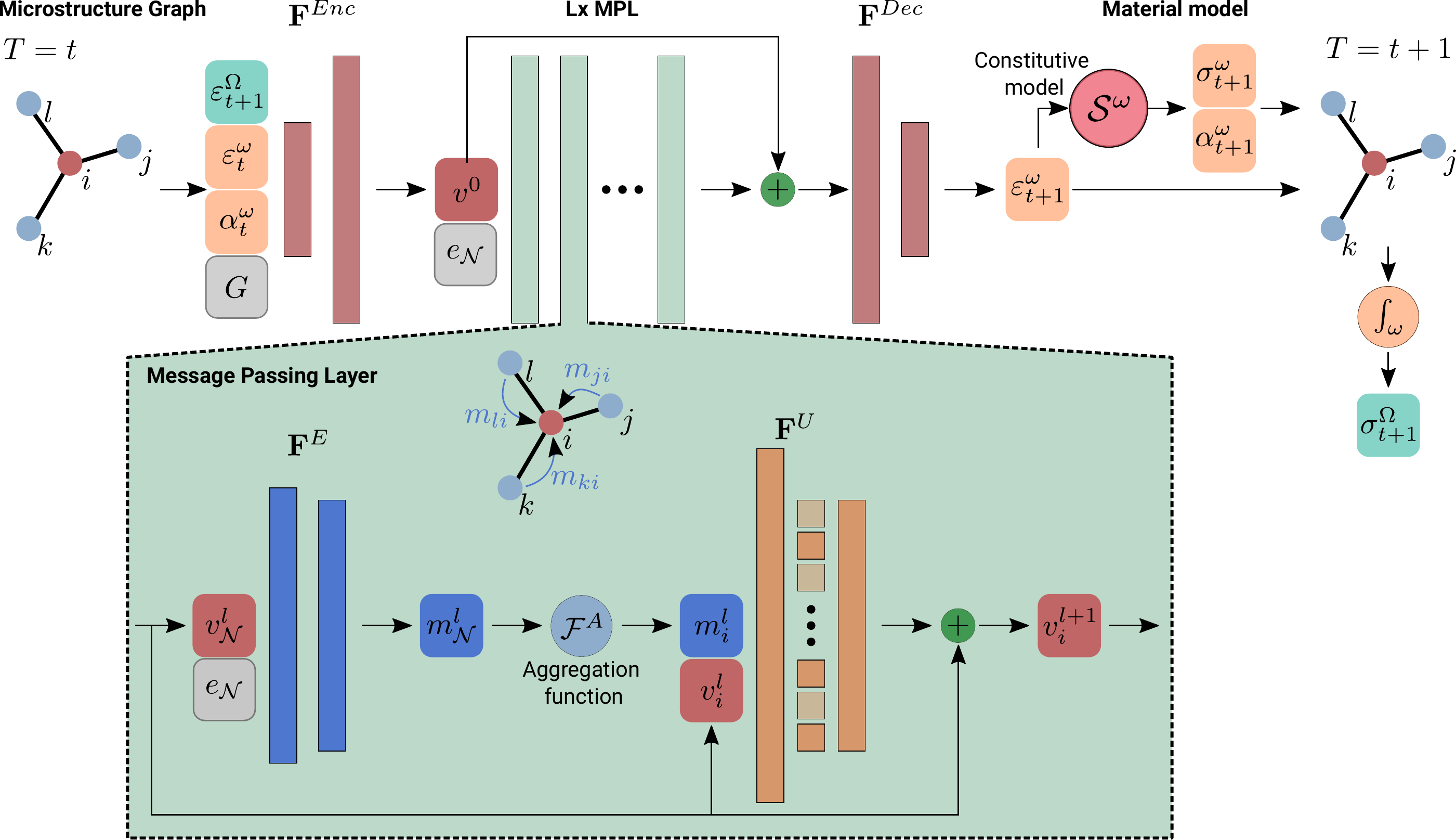}
    \caption{Visualization of the network architecture. The rectangles represent weight layers, where the height roughly indicates their number of weights. The center layer in $F^U$ represents dropout. }
    \label{fig:Network_architecture}
\end{figure}

Based on the graph and the node and edge features described in the previous section, the GNN predicts the strain for the next step $\boldsymbol{\varepsilon}^{\omega}_{t+1}$.
The predicted strains are then passed to a purely physics-based constitutive material model $\mathbf{\mathcal{S}}^{\omega}$ to compute the stresses $\boldsymbol{\sigma}^{\omega}_{t+1}$, this is the same constitutive model as used in the original micromodel.
In this process, the internal variables $\boldsymbol{\alpha}^{\omega}_{t+1}$ are also implicitly obtained.
In the multiscale setting, $\boldsymbol{\sigma}^{\omega}_{t+1}$ is homogenized using Eq. (\ref{eq:homog}) to obtain $\boldsymbol{\sigma}^{\Omega}_{t+1}$.
The updated strains $\boldsymbol{\varepsilon}^{\omega}_{t+1}$, internal variables $\boldsymbol{\alpha}^{\omega}_{t+1}$ and new macroscopic strain input $\boldsymbol{\varepsilon}^{\Omega}_{t+2}$ are then used in the next time step.
We present an abstract representation of the complete framework in Figure~\ref{fig:abstractGNN}.
Unlike other GNN-based models that directly predict all stress and displacement-related quantities as network outputs~\citep{pfaff2020learning,gladstone2023gnn}, our proposed method thus retains a physics-based material model.
We essentially extend an Encode-Process-Decode architecture~\cite{battaglia2018relational} to an Encode-Process-Decode-Material architecture, visualized in Figure~\ref{fig:Network_architecture}.
We believe that this offers an advantage by reducing the complexity of the problem to be learned.

We compute the messages as
\begin{equation}\label{eq:edge_update_used}
    \mathbf{m}_{ij} = F^E ( \mathbf{e}_{ij}, \mathbf{v}_j ),
\end{equation}
concatenating the neighboring nodal state with the edge features, and excluding the state of the current node.
We then aggregate the messages by summing them
\begin{equation}\label{eq:aggregation_used}
    \mathbf{m}_{i} = \sum_{j \in \mathcal{N}_i}^{} \mathbf{m}_{ij},
\end{equation}
where $\mathcal{N}_i$ represents the set of neighboring nodes of node $i$.
We concatenate the aggregated message with the current nodal state and pass this through the update function in Eq. (\ref{eq:update}) to obtain the new nodal state.
The encoder $F^{Enc}$, decoder $F^{Dec}$, edge message $F^E$, and node update $F^U$ networks are all chosen as multi-layer perceptrons with 2 layers each.
In the literature, we observe many slightly varying strategies for using dropout in the GNN architecture.
In~\citep{krokos2022graph}, dropout is applied after every fully connected layer,~\citep{gladstone2023gnn} between each MPL block, and in~\citep{vlassis2020geometric} between every GCN and Dense layer.
In our model, we found only minor differences in performance between these strategies, and have chosen to apply dropout inside each update function.
We use unique layers in each of the MPLs, as we found this to be more effective than sharing parameters between the layers.
Additionally, residual connections are added inside each MPL and from the encoded state to the final MPL output.

\subsection{Training process}
The aim of training the model is to learn the set of parameters of $F^{Enc}$, $F^{Dec}$, $F^E$ and $F^U$ that minimizes a loss function $\mathcal{L}$ on the data.
The direct output of the GNN-based network is $\varepsilon^{\omega}$, based on which we compute $\sigma^{\omega}$ through the material model, which we then homogenize to $\sigma^{\Omega}$.
We introduce the loss function based on these microscopic quantities as:
\begin{equation}\label{eq:xi}
    \mathcal{L}= \xi \mathcal{L}_{\sigma^{\omega}} + \left(1 - \xi\right) \mathcal{L}_{\varepsilon^{\omega}},
\end{equation}
where $\xi \in [0, 1]$ is a hyperparameter that allows us to emphasize either the full-field stress or full-field strain.
While $\mathcal{L}_{\sigma^{\omega}}$ and $\mathcal{L}_{\varepsilon^{\omega}}$ are directly related, we have found it crucial to include both terms in the loss function.
If only $\mathcal{L}_{\varepsilon^{\omega}}$ is considered, small errors for low-value strains can lead to large stress errors.
On the other hand, if only $\mathcal{L}_{\sigma^{\omega}}$ is considered, stresses can potentially still be correct even from considerably wrong strains, e.g if the micromodel exhibits perfectly plastic behavior.
This is caused by the nonlinearity of the material model, for a linear elastic microscale material this would not be necessary.
Their loss terms are computed as:
\begin{equation}\label{eq:loss_strainfield}
    \mathcal{L}_{\varepsilon^{\omega}}= \sqrt{ \dfrac{1}{N} \sum_{n=1}^{N} \dfrac{1}{T_n E_n C} \sum_{t=1}^{T_n} \sum_{e=1}^{E_n} \sum_{c=1}^{C} ( \hat{ \varepsilon }^{\omega, t}_{n,e,c} - \varepsilon^{\omega, t}_{n,e,c} )^2 },
\end{equation}
\begin{equation}\label{eq:loss_stressfield}
    \mathcal{L}_{\sigma^{\omega}}= \sqrt{ \dfrac{1}{N} \sum_{n=1}^{N} \dfrac{1}{T_n E_n C} \sum_{t=1}^{T_n} \sum_{e=1}^{E_n} \sum_{c=1}^{C} ( \hat{ \sigma }^{\omega, t}_{n,e,c} - \sigma^{\omega, t}_{n,e,c} )^2 }.
\end{equation}
Here we use $N$ as the number of samples, $T_n$ as the number of time steps, $E_n$ as the number of elements in that sample, and C represents the number of components, which is 3 in the 2D case.
In addition to these losses, we introduce the loss of $\sigma^{\Omega}$ which is the relevant quantity for the macroscale model in a multiscale simulation:
\begin{equation}\label{eq:loss_stresshom}
    \mathcal{L}_{\sigma^{\Omega}}= \sqrt{ \dfrac{1}{N} \sum_{i=1}^{N} \dfrac{1}{T_n C} \sum_{t=1}^{T_n} \sum_{c=1}^{C} ( \hat{\sigma}^{\Omega, t}_{n,c} - \sigma^{\Omega, t}_{n,c} )^2 }.
\end{equation}
This loss completely depends on $\sigma^{\omega}$, and we empirically find it unnecessary to include in the training loss.

Updating the network parameters requires backpropagating the loss $\mathcal{L}$ through all time steps included for the samples.
Because the input $\boldsymbol{\alpha}^{\omega}$ depends on the output of the previous step, and since we also include stress-based targets, the loss needs to be backpropagated through the material model.
The constitutive model $\mathbf{\mathcal{S}}^{\omega}$ implicitly evolves $\boldsymbol{\alpha}^{\omega}$, creating a recurrency.
Because of this, the loss at any time step depends on the predictions made for all previous time steps.
This leads to a complex computational graph and expensive gradient computations, which scale with the number of time steps during training.
An alternative approach is predicting only one time step ahead, but this can lead to unstable rollouts despite low training errors.
By artificially inducing noise on the inputs this discrepancy can be reduced~\citep{pfaff2020learning,brandstetter2022message}.
However, we find this one-step-ahead approach unnecessary here and instead directly propagate all time steps for a sample during training.

\subsection{Implementation}
The data was generated with an in-house Finite Element software developed using the Jem/Jive~\citep{nguyen2020jive} open-source C++ library.
The microstructures are represented as periodic 2D meshes generated using Gmsh~\citep{gmsh}.
The proposed GNN-based model was implemented using Pytorch Geometric~\citep{Fey/Lenssen/2019}.
The GNN-based model evaluates all material points in a batch in parallel, and the loss is propagated through them using automatic differentiation.
Training and time comparisons for the surrogate are performed on an NVIDIA Tesla V100S 32GB GPU, and FE simulations on an Intel Xeon E5-6248R 24C 3.0GHz, both in the DelftBlue cluster~\cite{DHPC2022}.
We provide code and a dataset to reproduce the results in this paper at \url{https://github.com/JoepStorm/Microscale-GNN-Surrogate}.

\section{Results - monotonic elasto-plasticity}\label{sec:monotonic}
We start the analysis by subjecting the micromodel to monotonically increasing strains.
To accurately reproduce these predictions, the GNN model needs to demonstrate the ability to predict in an auto-regressive manner for multiple time steps and account for plasticity for a wide variety of microstructures.
We discuss the relevance of hyperparameters and perform a model selection study before showing the results for the optimal model.
A limitation of the model when only trained on monotonic strain paths is highlighted, motivating the need for training on non-monotonic data, which we discuss in the next section.

\subsection{Data generation}
\begin{figure}[tb]
\centering
\includegraphics[width=1.0\textwidth]{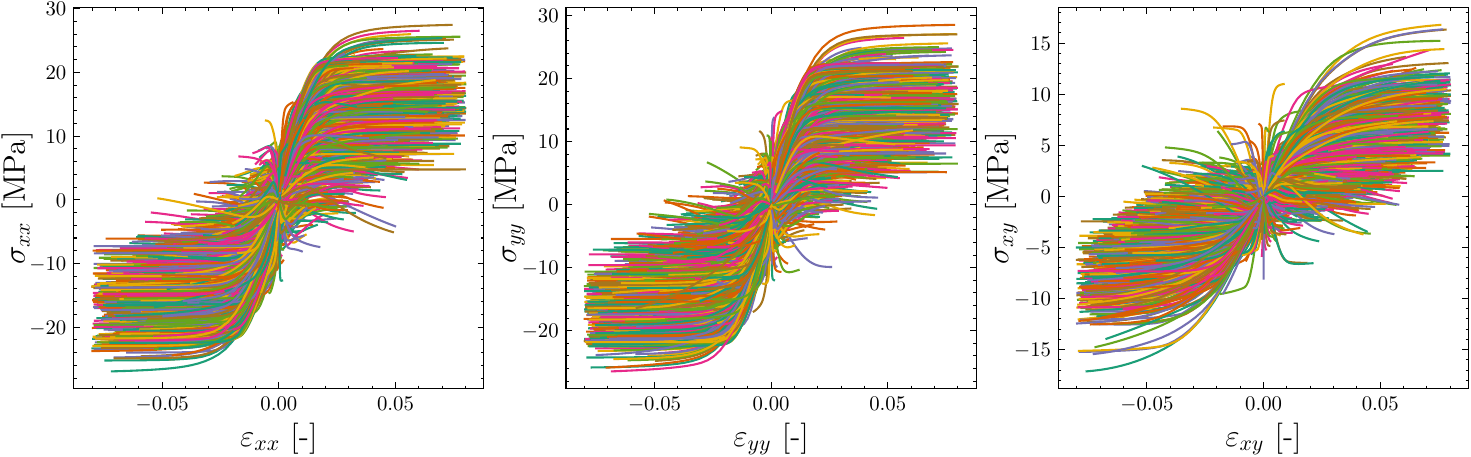}
\caption{Examples of stress-strain curves obtained from simulating one to nine void microstructures with a monotonic strain increase.}
\label{fig:load_curves_mono}
\end{figure}

We generate a different periodic microstructure for each sample (we define a sample as one complete strain path), with a random distribution of one to nine voids (each chosen an equal number of times) but keeping a fixed volume fraction $V_f = 0.6$.
We also sample a random average element size for each mesh, allowing the model to handle a wide range of mesh densities during inference.
Then, for each sample, a stress-strain path is generated by selecting a random strain direction and monotonically increasing the strain in this direction, keeping the loading proportional between components.
We store the macroscopic and microscopic quantities for each converged time step and discard intermediate Newton-Raphson iterations.
We use an elasto-plastic material with von Mises plasticity (J2), a Young's modulus of $3130 \; [MPa]$, a Poisson's ratio of $0.37 \; [-]$, and yield criterion $\sigma_C = 64.80 - 33.60 \cdot e^{\frac{\varepsilon^p_{eq}}{-0.003407}}$, where $\varepsilon^p_{eq}$ is the equivalent plastic strain (the internal variable $\boldsymbol{\alpha}^{\omega}$ of this material model).
In Figure~\ref{fig:load_curves_mono} we show examples of the resulting macroscopic stress-strain curves after stress homogenization.
We generate 4000 samples as training data, 2000 as a validation set, and additional test sets also consist of 2000 samples unless specified otherwise.
In section~\ref{sec:nonmono}, we will show that 4000 training samples are sufficient by plotting the learning curve for the training sample size for a more complex case.

\subsection{Model selection}\label{sec:monotonic_model_selection}
For any deep learning architecture, there are many variables that have a considerable influence on the model performance.
Specifically for GNNs, these include the number of MPLs, the number of neurons per layer, the number of layers per MPL, the number and width of layers of the encoder and decoder, the activation functions, normalization layers, residual connections, the aggregation function, dropout layers and their rate, the optimizer, the learning rate, and the batch size.
In addition, many of these hyperparameters are interdependent and have many possible values.
Performing a thorough model selection study is thus crucial for each problem setting.
However, due to the vast number of options and the computational cost of training the model, it is not feasible to consider all combinations.
Therefore, a practical approach is necessary for determining the optimal hyperparameters.

Some settings are chosen based on previous work, such as employing two layers per multi-layer perceptron as chosen in MeshGraphNets~\citep{pfaff2020learning}.
For other architectural choices, such as choosing if and where to add normalization layers and residual connections, we compare several options during the initial model creation and then keep this architecture fixed for the rest of the study as described in section~\ref{sec:architecture}.
In purely data-driven methods the parameter count, often determined by the amount of neurons per layer, has a significant influence on the inference and training speed of the network.
In our case, the computational time is primarily influenced by the material model instead (and scales only with the number of elements).
Therefore we can afford a very high parameter count without significantly compromising in computation speed, giving us a highly flexible model, and use the dropout rate as a hyperparameter to regularize the model.
The high parameter count follows from using 512 neurons per layer, which is then also the dimensionality of the node state $\mathbf{v}_i$ and messages $\mathbf{m}_i$.
For the learning rate, we use 10 warmup epochs to go from $1e^{-5}$ to $2e^{-4}$ and follow this up with an exponential decay with factor $0.998$: $2e^{-4} \cdot 0.998^{epoch}$ for a fixed time budget of 105 hours giving around 700 epochs for a typical model.
The number of considered void features is kept at 9, the maximum number of voids included in training samples.

\begin{figure}[htbp]
    \centering
    \begin{subfigure}[t]{0.32\textwidth}
      \centering
      \includegraphics[width=1\textwidth]{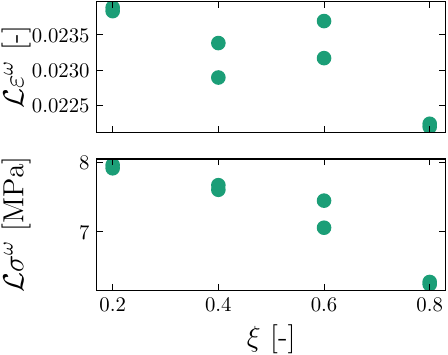}
      \caption{Round 1: $\xi$ value}
      \label{fig:xi_1}
    \end{subfigure}%
    \hfill
    \begin{subfigure}[t]{0.32\textwidth}
      \centering
      \includegraphics[width=1\textwidth]{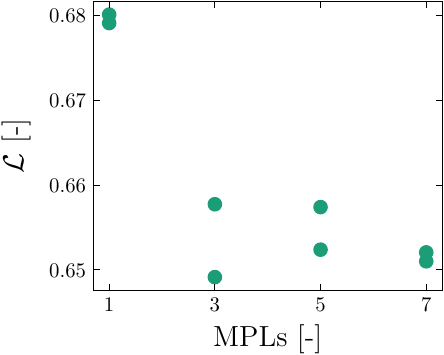}
      \caption{Round 1: Number of MPLs}
      \label{fig:MPLs_1}
    \end{subfigure}%
    \hfill
    \begin{subfigure}[t]{0.32\textwidth}
      \centering
      \includegraphics[width=1\textwidth]{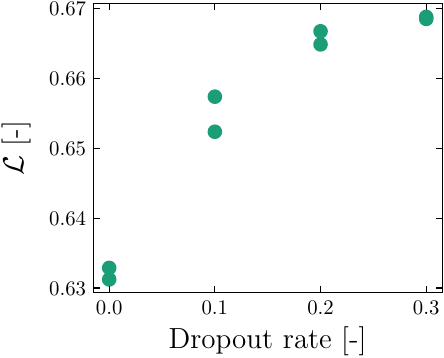}
      \caption{Round 1: Dropout rate}
      \label{fig:drop_1}
    \end{subfigure}%
    \hfill
    \begin{subfigure}[t]{0.32\textwidth}
      \centering
      \includegraphics[width=1\textwidth]{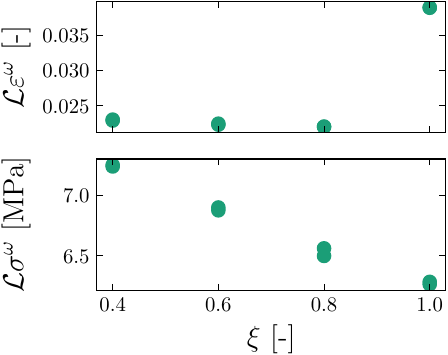}
      \caption{Round 2: $\xi$ value}
      \label{fig:xi_2}
    \end{subfigure}%
    \hfill
    \begin{subfigure}[t]{0.32\textwidth}
      \centering
      \includegraphics[width=1\textwidth]{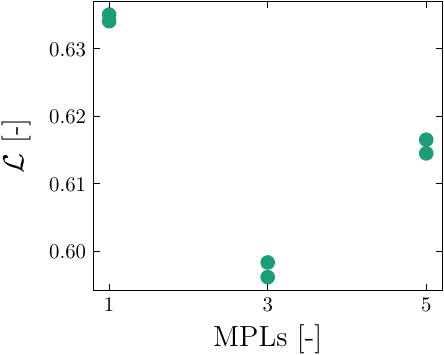}
      \caption{Round 2: Number of MPLs}
      \label{fig:MPLs_2}
    \end{subfigure}%
    \hfill
    \begin{subfigure}[t]{0.32\textwidth}
      \centering
      \includegraphics[width=1\textwidth]{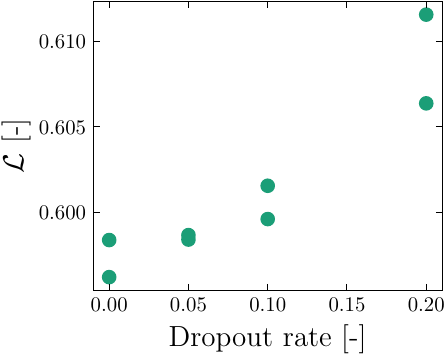}
      \caption{Round 2: Dropout rate}
      \label{fig:drop_2}
    \end{subfigure}%
  \caption{The results for two rounds of the hyperparameter study. The initial values chosen as default values for round 1 are $\xi=0.6$, MPLs=5, and dropout=0.1.
  For $\xi$ we plot the individual (unnormalized) validation losses $\mathcal{L}_{\varepsilon^{\omega}}$, $\mathcal{L}_{\sigma^{\omega}}$ separately, as their combined loss is itself dependent on $\xi$.
  Based on round 1, we choose the default values for round 2 as $\xi=0.8$, MPLs=3, and dropout=0.0.}
  \label{fig:hyperstudy_mono}
\end{figure}

MPLs are the defining feature of GNNs, and the number of MPLs is an important hyperparameter we study.
As we are dealing with elasto-plasticity, we find that the tradeoff between strains and stresses using $\xi$ (Eq.~\ref{eq:xi}) is also an important hyperparameter to consider.
For the number of MPLs, the dropout rate, and $\xi$ we perform several rounds of model selection, where in each round we vary one parameter at a time and use the best value from the previous round as a starting point until the values converge.
We train two models per setting, and the resulting errors for each variable in these rounds are shown in Figure~\ref{fig:hyperstudy_mono}.
In both rounds, mainly the value of $\xi$ and the dropout rate have an influence.
The small influence of the number of MPLs is surprising and should be investigated further in future work.

Choosing the value for $\xi$ is not straightforward, as the nature of the optimization problem changes as we change the objective by adjusting $\xi$.
Considering only the value of the combined validation loss is therefore insufficient.
Instead, we need to consider the components $\mathcal{L}_{\varepsilon^{\omega}}$ and $\mathcal{L}_{\sigma^{\omega}}$ separately.
Based on these two separate losses, we conclude that a value of $\xi=0.8$ is a good choice for our problem.
Remarkably, we observe that having a value of $0 < \xi < 0.8$ not only decreases $\mathcal{L}_{\sigma^{\omega}}$ but does so without compromising $\mathcal{L}_{\varepsilon^{\omega}}$.

Since the best settings after the second round are the same as used as the base values of the second round, we do not need to perform an additional round.
We can use the model with 3 MPLs, $\xi=0.8$, and a dropout rate of 0.0 for further experiments.

\subsection{Prediction results}
\begin{figure}[htbp]
\centering
\includegraphics[width=1.\textwidth]{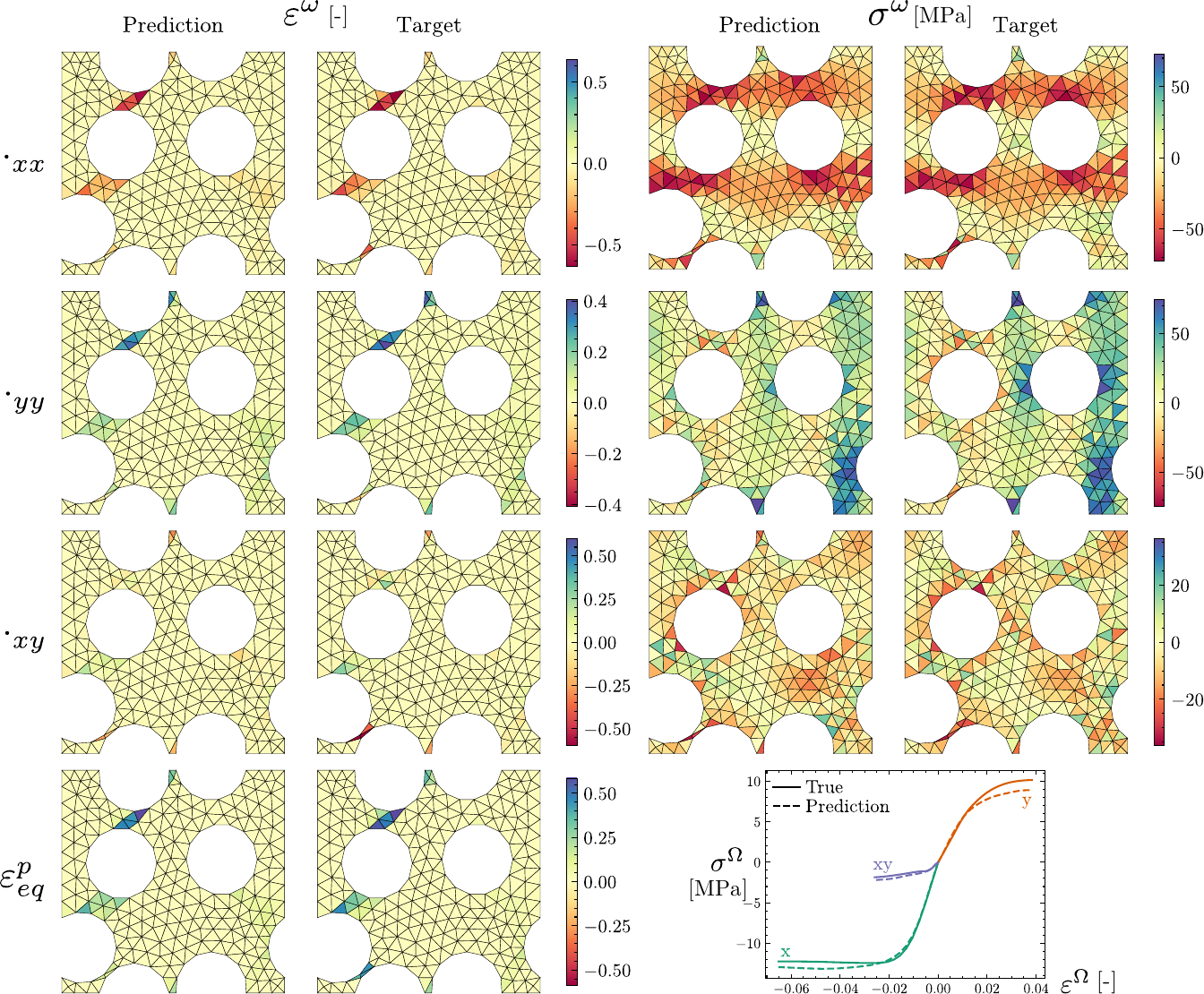}
\caption{Example of all full-field predictions made after 25 time steps and the complete homogenized stress-strain curve. Due to the elasto-plastic material the strain and stress fields have different patterns - the strain fields are dominated by a few elements with large plastic deformation.}
\label{fig:mono_example_res}
\end{figure}

With the model determined, we can now evaluate the prediction results.
Figure~\ref{fig:mono_example_res} shows all microscopic quantities for a 5-void test sample subjected to a random strain direction after 25 time steps and the resulting macroscopic stress-strain curve, all in comparison with target results from the full micromodel.
This example also demonstrates the differences in the strain and stress fields as a result of the nonlinear material model.
Even after 25 steps, the surrogate model correctly predicts the areas of high and low strain, and this is well reflected in the equivalent plastic strain $\varepsilon^p_{eq}$ field (computed by the physics-based material models $\mathbf{\mathcal{A}}^{\Omega}$ at each integration point).
Furthermore, the image highlights the difficulty of the problem to solve: the $\sigma^{\omega}_{xy}$ field has little recognizable structure, yet the model can predict it well.

\subsection{Extrapolating for non-monotonic paths}\label{sec:mono_unload}
A conventional neural network surrogate, trained solely on the macroscopic strain $\varepsilon^{\Omega}$, cannot differentiate between loading and unloading, resulting in an unloading path identical to the loading path.
However, the model has two additional sources of information that could theoretically aid in distinguishing between loading and unloading.
Firstly, the model is provided not only with the next macroscopic strain but also with the current microscopic strain and history variables.
Secondly, the embedded material model implicitly tracks its history, and recent surrogate models with embedded material models (albeit in a very different manner) have been shown to be able to predict unloading without having been trained on it~\citep{maia2022physically, liu2019deep}.

\begin{figure}[tb]
\centering
\begin{minipage}[t]{.45\textwidth}
    \centering
    \includegraphics[width=.9\textwidth]{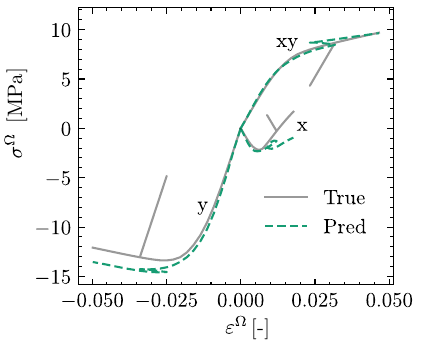}
    \caption{The behavior of a GNN trained on monotonically increasing data, predicting a load case with unloading.}
    \label{fig:mono_unload}
\end{minipage}
\hfill
\begin{minipage}[t]{.45\textwidth}
    \centering
    \includegraphics[width=.8\textwidth]{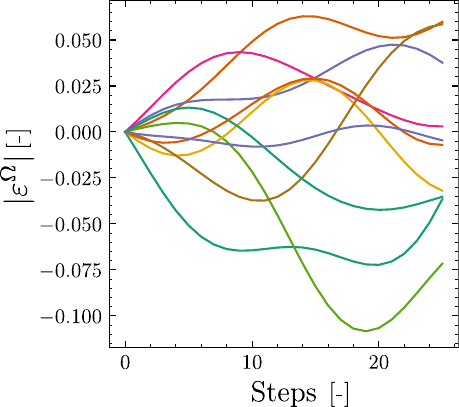}
    \caption{The samples of the GP used for the strain magnitude.}
    \label{fig:gp_samples}
\end{minipage}
\end{figure}

To explore the potential of the model, we examine a scenario involving unloading.
We create a new test set, but instead of monotonically increasing the strain, we decrease it from steps 15 to 19, unloading the microstructure, before resuming loading from steps 20 to 30 (keeping the magnitude of the strain increment fixed).
In Figure~\ref{fig:mono_unload} we show the results for a sample and observe that the model fails to capture unloading, as during the unloading phase the stress predictions are far away from the true unloading branch.
Despite the good predictions for monotonic paths, the model does not implicitly learn to predict unloading from monotonic strain paths.
In the following section, we therefore train it on non-monotonic data.

\section{Non-monotonic elasto-plasticity}\label{sec:nonmono}
In an FE analysis, the local strain evolution in individual elements can be highly nonlinear functions of the evolution of boundary conditions.
In the previous section, we demonstrated that the model can make accurate predictions for monotonically increasing load paths with elasto-plastic behavior, but it is not yet able to predict the unloading response.
Therefore, we now increase the complexity of the training data, to improve prediction results for more general load cases.

\subsection{Data generation}
\begin{figure}[tb]
\centering
\includegraphics[width=1.0\textwidth]{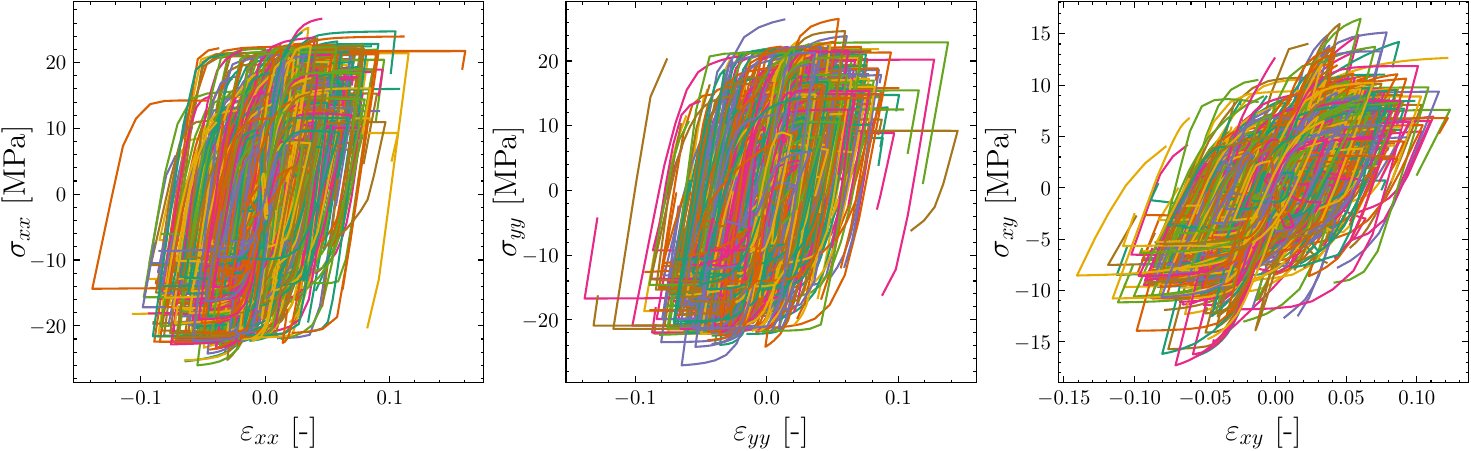}
\caption{Examples of stress-strain curves obtained from simulating 1 to 9-void microstructures following a non-monotonic, GP-based strain magnitude.}
\label{fig:load_curves_gp}
\end{figure}

To create arbitrary load paths, we generate random walks based on Gaussian Processes (GP), as done in several other works~\citep{mozaffar2019deep, logarzo2021smart, maia2022physically}.
We retain a fixed random strain direction and proportional loading, but the norm of the strain vector now follows the magnitude of a GP sample.
Specifically, we initialize a squared exponential GP with a variance of $2e^{-3}$, a length scale of 8, and fit through the point [0,0].
We draw random realizations from this GP for 25 time steps per realization; ten of these are shown in Figure~\ref{fig:gp_samples}.
The generation of the microstructures and the material properties are the same as for the monotonic loading case.
In Figure~\ref{fig:load_curves_gp} the resulting ground-truth stress-strain curves are shown.

\subsection{Model selection}
\begin{figure}[htbp]
    \centering
    \begin{subfigure}[t]{0.32\textwidth}
      \centering
      \includegraphics[width=1\textwidth]{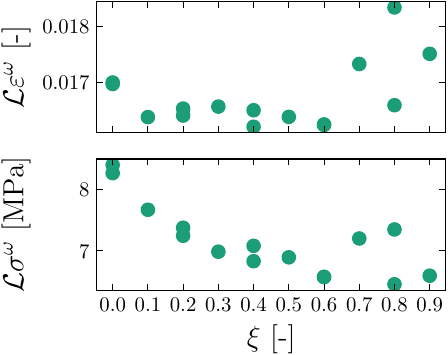}
      \caption{Round 1: $\xi$ value}
      \label{fig:nonmono_xi_1}
    \end{subfigure}
    \hfill
    \begin{subfigure}[t]{0.32\textwidth}
      \centering
      \includegraphics[width=1\textwidth]{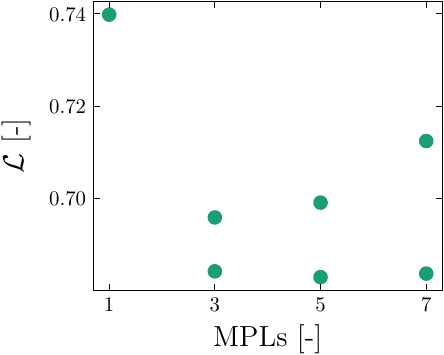}
      \caption{Round 1: Number of MPLs}
      \label{fig:nonmono_MPLs_1}
    \end{subfigure}
    \hfill
    \begin{subfigure}[t]{0.32\textwidth}
      \centering
      \includegraphics[width=1\textwidth]{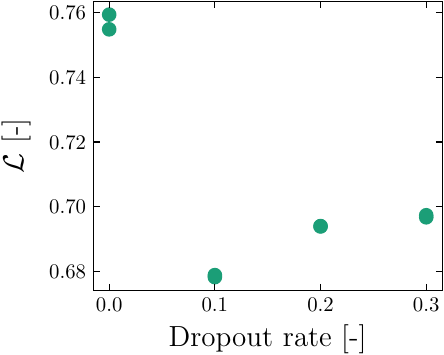}
      \caption{Round 1: Dropout rate}
      \label{fig:nonmono_drop_1}
    \end{subfigure}
  \caption{The results for the hyperparameter study. The initial values chosen as default values for round 1 are $\xi$=0.6, MPLs=5, and dropout=0.1. Values for $\xi=1$ are omitted from the plots as they lead to a very high $\mathcal{L}\varepsilon^{\omega}$. Since the optimal values are the same as the base values, we do not perform a second round.}
  \label{fig:hyperstudy_nonmono}
\end{figure}

We perform the same model selection process as in the section~\ref{sec:monotonic_model_selection}, but now for models trained with the GP-based data.
The same initial values are used for the first round, namely $\xi=0.6$, 5 MPLs, and a dropout rate of 0.1, and we vary each of these parameters one at a time.
The losses are computed on a validation set of 2000 GP-based samples, therefore these losses cannot be compared directly to those of the monotonic case.
We visualize the results in Figure~\ref{fig:hyperstudy_nonmono}.
From this first round of model selection, we find that the optimal model is obtained with $\xi=0.6$, 5 MPLs, and a dropout rate of 0.1.
Again we find little difference between the errors in the number of MPLs.
As these values for the hyperparameters are the same as the initial values, no second round is performed, and the model with these settings will be used for all following experiments.

\begin{figure}[htbp]
\centering
\includegraphics[width=.4\textwidth]{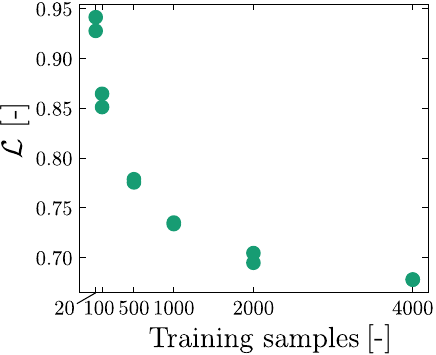}
\caption{The learning curve for models trained based on GP-based load paths, indicating that using 4000 training samples is sufficient. For each setting of training samples, we show the result of two training runs with random initialization.}
\label{fig:learncurve}
\end{figure}

So far we have been assuming that 4000 training samples are sufficient for the model and that providing more data would not significantly improve the model accuracy.
Here we show that this is indeed the case using a learning curve by training models with a varying number of training samples.
When the model accuracy does not increase significantly with more data, we take it as an indication the dataset size is sufficiently large.
We show the test error of the various models in Figure~\ref{fig:learncurve}.
The trend has not completely converged and a lower error could possibly be obtained with more samples, but this also comes at the cost of requiring more expensive data.
We therefore consider that using 4000 samples is sufficient.
As this GP-based loading is more complex than monotonic loading, we argue that using the same amount of samples is more than enough for the monotonic case considered before.

\subsection{Prediction results}
\begin{figure}[htbp]
\centering
\begin{minipage}[t]{.45\textwidth}
  \centering
    \includegraphics[width=.9\textwidth]{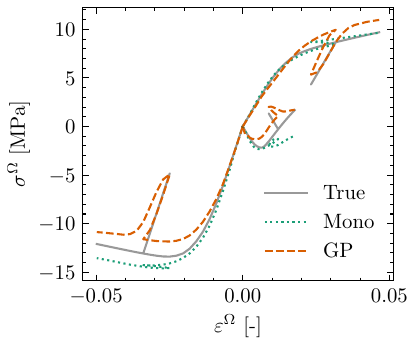}
    \caption{Comparison between a model trained on GP data and a model trained on monotonic data in predicting the homogenized stress-strain curve of an unloading test sample.}
    \label{fig:unloading_nonmono}
\end{minipage}
\hfill
\begin{minipage}[t]{.45\textwidth}
  \centering
    \includegraphics[width=.9\textwidth]{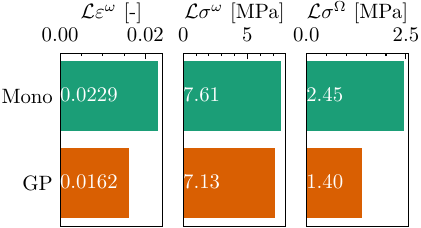}
    \caption{Comparison of the test errors for unloading between the models trained with monotonic and GP-based data.}
    \label{fig:unload_mono_vs_gp}
\end{minipage}
\end{figure}

Now, we return to the manually defined unloading case from Section~\ref{sec:mono_unload} and compare the performance of the model trained on the GP-based data to that of the model trained only on the monotonic data.
In Figure~\ref{fig:unloading_nonmono} we qualitatively show the resulting $\boldsymbol{\varepsilon}^{\Omega}-\boldsymbol{\sigma}^{\Omega}$ curves for both models.
The initial loading phase predictions are very similar for both models, but the model trained on the GP-based data can predict the unloading phase much better.
Still, predictions are far from perfect, and alternative surrogates only trained to predict $\sigma^{\Omega}$ can outperform this model.

In addition to these qualitative results, we also compare the test errors for the two models.
The results in Figure~\ref{fig:unload_mono_vs_gp} show an improved accuracy when training on the GP-based models.
Notably, there is a larger improvement in the loss associated with $\sigma^{\Omega}$ than with $\sigma^{\omega}$.
This indicates that while predicting which elements are subject to how much unloading is still challenging, the overall unloading pattern of the field is predicted much better.
It is clear from these experiments that training on non-monotonic data is necessary.
With a model able to make full-field predictions for complex load scenarios, we now want to better understand the influence of the physics-based material model retained by the surrogate.

\subsection{Material model}
One of our main contributions is including a purely physics-based material model inside the surrogate to model nonlinear material behavior.
In this section, we investigate the influence of the added material model on surrogate performance.
The material model provides two additions, which we attempt to study separately as best as possible.
First, we consider a model for which we remove the input feature of the equivalent plastic strain $\varepsilon^p_{eq}$, which is obtained when computing local stresses.
Second, we study the influence of directly computing the stresses by removing the material model and instead predicting the stresses as GNN outputs while keeping the loss function the same.
This data-driven option without the material model is closest to most GNN-based surrogate alternatives, and as it increases the complexity of the model, we expect it to require more training data to learn.
We compare one additional model where we change the nature of the problem by training only for strains but still using the material model to compute stresses during inference.
We achieve this by setting $\xi=0$, and since we do not compute the material model during training, we also do not use the $\varepsilon^p_{eq}$ feature.
We thus consider three alternatives to the base model, which we summarize in Table~\ref{tab:matmodels}.

\newcolumntype{C}{>{\centering\arraybackslash} m{2.5cm} }  
\begin{center}
\begin{table}[!htbp]
    \centering
    \caption{Overview of alternative models.}
    \begin{tabular}{|m{2.5cm}|C|C|c|c|}
        \hline
        Model & Material model during training & Material model during inference & $\varepsilon^p_{eq}$ feature & $\xi$ \\
        \hline
        \hline
        A: Base model & \cmark & \cmark & \cmark & 0.6 \\
        B: No $\varepsilon^p_{eq}$ & \cmark & \cmark & \xmark & 0.6 \\
        C: No material & \xmark & \xmark & \xmark & 0.6 \\
        D: Strain-based & \xmark & \cmark & \xmark & 0.0 \\
        \hline
    \end{tabular}
    \label{tab:matmodels}
\end{table}
\end{center}

\begin{figure}[tb]
\centering
\includegraphics[width=1\textwidth]{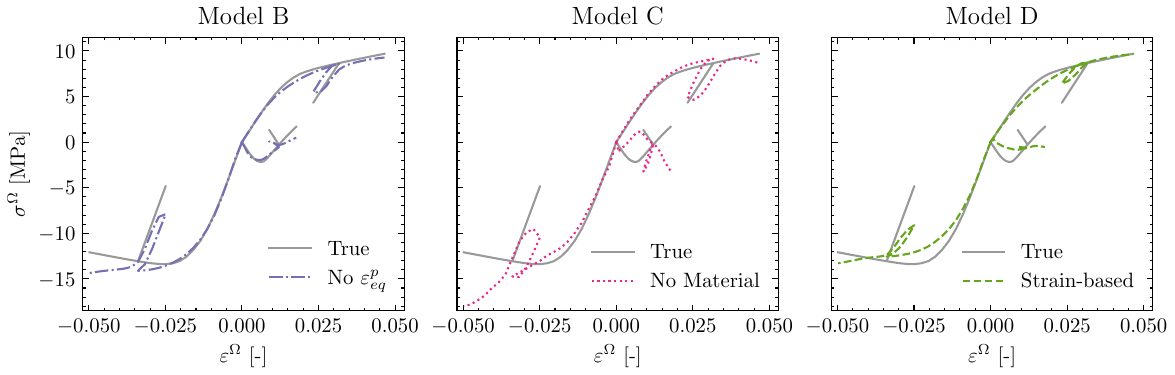}
\caption{A comparison between homogenized stress-strain curve predictions of the three alternative models (cf. Figure~\ref{fig:unloading_nonmono} for model A).}
\label{fig:unloading_nomat}
\end{figure}

In Figure~\ref{fig:unloading_nomat} we compare a $\varepsilon^{\Omega} - \sigma^{\Omega}$ curve between the different models.
Since model B performs similarly to the base model (shown in Figure~\ref{fig:unloading_nonmono}), it is unclear whether the $\varepsilon^p_{eq}$ feature is beneficial, at least for the current architecture.
We observe that model C, without the material model, predicts the stresses well during loading but cannot predict the unloading phase.
In this example, model D already shows a large error in the monotonic phase and does not predict the unloading slope well either.

\begin{figure}[htbp]
\centering
\includegraphics[width=0.7\textwidth]{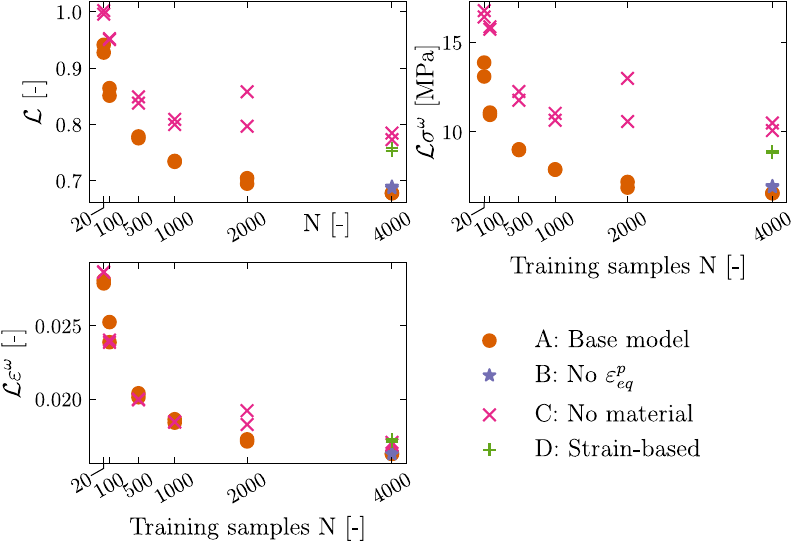}
\caption{The learning curves on a GP-based test set for the various alternative models. The individual components show the unnormalized losses.}
\label{fig:learncurve_mat}
\end{figure}

We also quantitatively compare the models on two different test sets, one with GP-based paths, and the other is the unloading test set introduced in Section~\ref{sec:mono_unload}.
Starting with the GP-based test set, we compare the learning curves of the different models in Figure~\ref{fig:learncurve_mat}.
The error is significantly lower for model A than for model C, regardless of the number of training samples.
This indicates that the material model directly results in better predictions rather than increasing the rate at which it learns.
The improvement in overall accuracy predominantly originates from the stress predictions.

\begin{figure}[htbp]
\centering
\includegraphics[width=.45\textwidth]{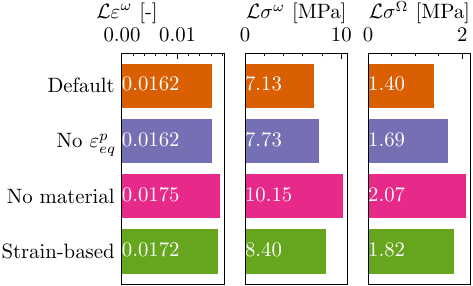}
\caption{A comparison of the test errors for unloading between the models with various alternatives to the material model.}
\label{fig:unload_matbarchart}
\end{figure}

Model B produces very similar results to model A for the GP-based dataset, indicating that the $\varepsilon^p_{eq}$ feature only represents a minor contribution to performance.
We show the errors for the unloading test set in Figure~\ref{fig:unload_matbarchart}, where we observe a slightly increased difference between models A and B.
Still, we expected the model to rely more on this feature to accurately predict the stresses, reflecting what happens in the actual FE micromodel.
Across both test sets, the losses of models C and D are higher than the base model, indicating that there is a significant benefit from including the material model.
Model D outperforms model C, indicating that training with the stresses directly as a model output can be detrimental to the performance compared to only computing them with the actual material model during inference.
We observe that the strain errors are similar between all models, and the difference in the error is mainly due to the stress predictions.

The contribution of the material model does come at a significant computational cost.
To start, the history dependency of the material model complicates the computational graph for backpropagation.
But more significant is the computational cost of evaluating the material model itself.
As the material model is evaluated for each element in the mesh in the same way as when using the FE method, it scales with the number of elements and is therefore independent of the GNN hyperparameters.
This can become a significant computational bottleneck for expensive material models, as this operation needs to be performed many times during training.

\subsection{Predicting for more time steps}
\begin{figure}[htbp]
    \begin{minipage}[t]{0.66\linewidth}
        \centering
        \begin{subfigure}{0.5\textwidth}
          \centering
          \includegraphics[width=1\textwidth]{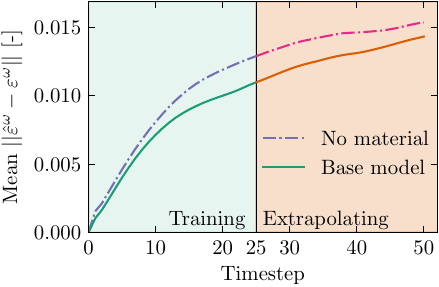}
          \caption{Average $\varepsilon^{\omega}$}
        \end{subfigure}%
        \hfill
        \begin{subfigure}{0.5\textwidth}
          \centering
          \includegraphics[width=1\textwidth]{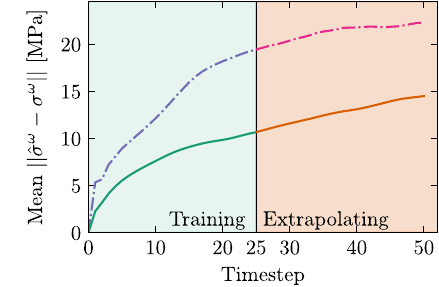}
          \caption{Average $\sigma^{\omega}$}
        \end{subfigure}%
        \hfill
        \begin{subfigure}{0.5\textwidth}
          \centering
          \includegraphics[width=1\textwidth]{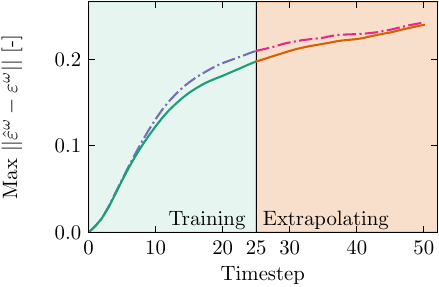}
          \caption{Max $\varepsilon^{\omega}$}
        \end{subfigure}%
        \hfill
        \begin{subfigure}{0.5\textwidth}
          \centering
          \includegraphics[width=1\textwidth]{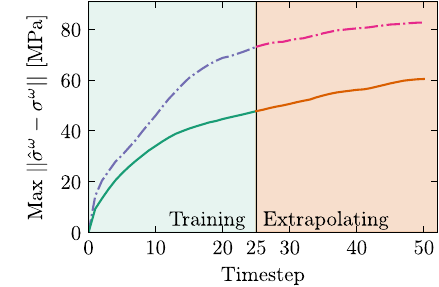}
          \caption{Max $\sigma^{\omega}$}
        \end{subfigure}%
    \end{minipage}
    \hfill
    \begin{minipage}[t]{0.33\linewidth}
        \vfill
        \begin{subfigure}{1\textwidth}
          \centering
          \includegraphics[width=1\textwidth]{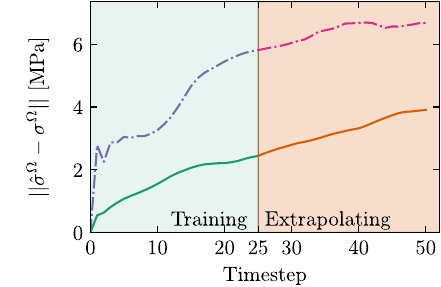}
          \caption{$\sigma^{\Omega}$}
        \end{subfigure}
        \vfill
    \end{minipage}
\caption{Average errors based on 2000 test samples, where we extrapolate beyond the number of time steps seen during training. For each sample, the errors are computed for each time step. In the top row we plot the average error over all samples, in the bottom row we compute the maximum error for each sample and time step, before averaging over these maxima. We plot the quantities for base model A and model C trained without the embedded material.}
\label{fig:t_extrap_plots}
\end{figure}

\begin{figure}[htbp]
    \centering
    \includegraphics[width=1.\textwidth]{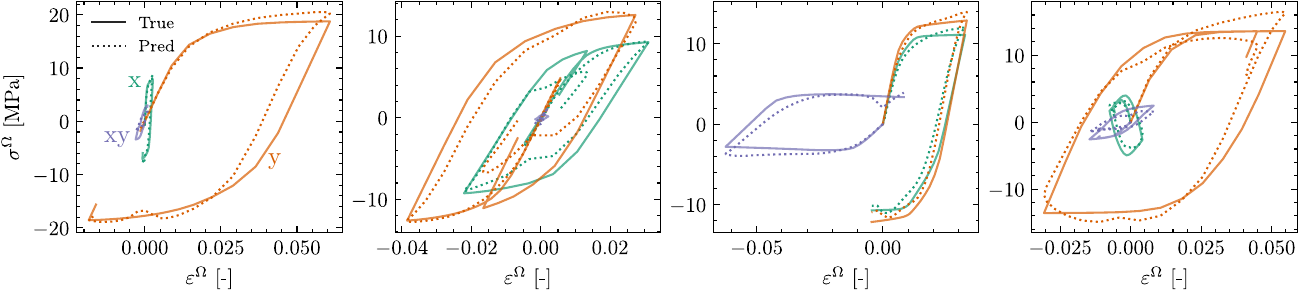}
    \caption{Examples of homogenized stress-strain curves for 50 time steps. }
    \label{fig:t_extrap_hom}
\end{figure}

In autoregressive models, errors can accumulate over time as their inputs depend on their previous outputs.
This can potentially lead to out-of-distribution inputs, which can cause exponentially increasing errors to occur.
We have only focused on the behavior of the model during the first 25 time steps, which was also the length of the training paths, and now investigate how the model behaves when extrapolating beyond this point.
We generate a new test set of 2000 load curves with 50 time steps.
For the model trained on samples with 25 time steps, we compute a mean and maximum error for each time step on this test set and average over all samples.
We compute the norm of the difference between the ground truth and predicted fields as errors that can be interpreted well:
\begin{equation}
    \text{Mean error} = \frac{1}{N} \sum_{n=1}^{N} \frac{1}{E_n} \sum_{e=1}^{E_n} ||  \boldsymbol{ \varepsilon}^{\omega}_{n, e} - \hat{ \boldsymbol{ \varepsilon }}^{\omega}_{n, e} ||,
\end{equation}
\begin{equation}
    \text{Max error} = \frac{1}{N} \sum_{n=1}^{N} \max( || \boldsymbol{ \varepsilon}^{\omega}_{n, e} - \hat{ \boldsymbol{ \varepsilon }}^{\omega}_{n, e} || \; \forall \; e \in E_n ).
\end{equation}
We compute similar errors for $\sigma^{\omega}$ and $\sigma^{\Omega}$ and compare the performance of the model A to model C (without a material model).
The results are shown in Figure~\ref{fig:t_extrap_plots}, from which we can make two main observations.
Firstly, after the errors increase quickly in the first few steps, the errors do not increase excessively as we go beyond the number of training steps but instead follow a near-linear trend.
Secondly, comparing the model to the one without a material model, we again observe that while the strain errors are similar, a large discrepancy exists for the stress errors.
Especially for the homogenized stress, the difference is already significant after the first few steps, where it fails to correctly predict the fast increase in stress during the elastic phase without the material model.
This difference then stays relatively constant as we extrapolate beyond the number of training steps.
The linear increase in error indicates that the model is stable when extrapolating to more time steps.
We show several representative examples in Figure~\ref{fig:t_extrap_hom}, where we can qualitatively observe that the models indeed still follow the correct trend.

\subsection{Microstructure scaling}\label{sec:rve_scaling}
\begin{figure}[htbp]
    \centering
    \includegraphics[width=.4\textwidth]{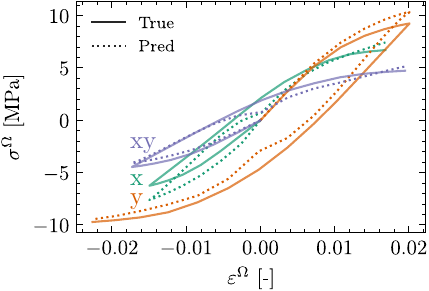}
    \caption{Homogenized stress-strain curve for a 49-void sample.}
    \label{fig:upscale_hom}
\end{figure}

\begin{figure}[htbp]
    \centering
    \includegraphics[width=1\textwidth]{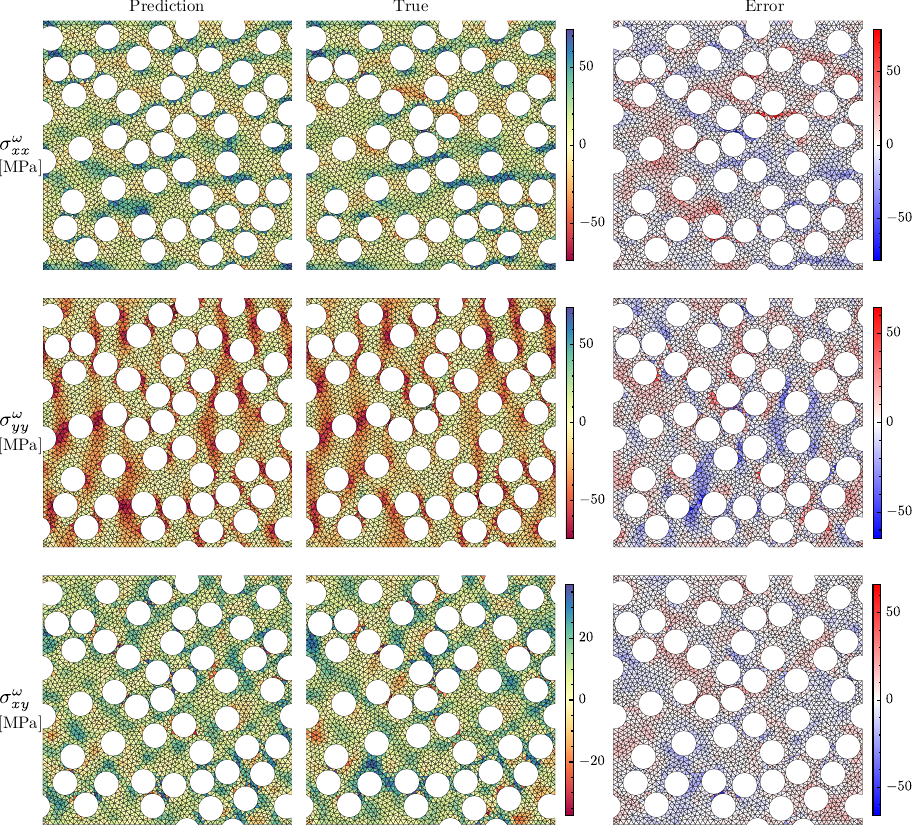}
    \caption{Full-field stresses after 25 time steps for a 49-void sample.}
    \label{fig:upscale_field}
\end{figure}

By training the model on various microstructures with a different number of voids, the model is encouraged to learn to predict general microstructures.
Here we investigate how well the model can extrapolate to larger microstructures.
We demonstrate the model predictions for a representative 49-void\footnote{As a reminder, training is performed with microstructures with no more than nine voids.} sample by plotting the homogenized stress-strain curve in Figure~\ref{fig:upscale_hom} and the full-field stresses in Figure~\ref{fig:upscale_field}.
While the microstructure size is never explicitly given to the model, the homogenized curves match very well.
We observe that the model can predict the general trend of the full-field stresses, although the predictions are not as accurate as for the smaller microstructures.
We do not plot the strains as they are dominated by a few elements with very high strains and therefore difficult to compare visually.

One quantity that depends on the microstructure size is the initial stiffness: for small microstructures, the stiffness is not representative of the real behavior.
When using the FE method, the homogenized stiffness is explicitly computed using the probing method.
For our model, we can compute the stiffness as
\begin{equation}
    \dfrac{\partial \sigma^{\Omega}}{\partial \varepsilon^{\Omega}} = \dfrac{\partial \sigma^{\Omega}}{\partial \sigma^{\omega}} \dfrac{\partial \sigma^{\omega}}{\partial \varepsilon^{\omega}}  \dfrac{\partial \varepsilon^{\omega}}{\partial \varepsilon^{\Omega}},
\end{equation}
which we perform using automatic differentiation.
In theory, this can be achieved by applying automatic differentiation after predicting with input $\varepsilon^{\Omega}=0$.
Since our GNN is a highly nonlinear model, using only $\varepsilon^{\Omega}=0$ could potentially give a distorted result, and we instead average over several very small random $\varepsilon^{\Omega}$ values.\footnote{In practice, we find the difference from doing this to be negligible.}

\begin{figure}[htbp]
\centering
\includegraphics[width=0.45\textwidth]{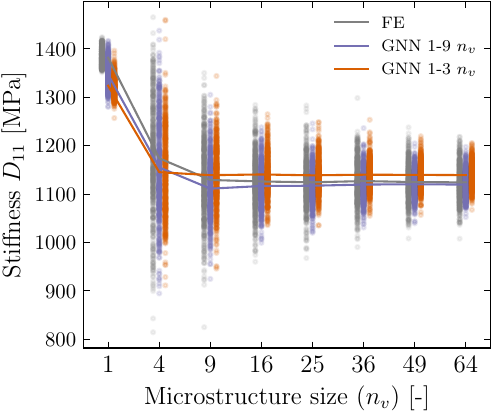}
\caption{A comparison of the stiffnesses obtained from the surrogate and an FE analysis for various microstructure sizes. The scatter shows 500 samples for each $n_v$, with the line giving the mean for each type. The surrogate models are trained on a limited number of voids, beyond which they are extrapolating. }
\label{fig:Transfer_stiffness}
\end{figure}

We create test datasets with varying microstructure sizes to test the ability to extrapolate to larger unseen geometries.
The size is determined by the number of voids in the microstructure, as the volume fraction and void size are kept constant.
For each size category, we generate 500 samples and compute the stiffness with an FE analysis and our model.
We present the resulting stiffnesses in Figure~\ref{fig:Transfer_stiffness}.
Our models, trained on data with an equal distribution of [1,2,3,4,5,6,7,8,9]-void microstructures, closely follow the FE result, for both the average stiffness but also the variance.
Since the stiffness is already almost converged after 9 voids, which the model has seen during training, we train another model where we only include [1,2,3]-void microstructures during training.
This model still follows the trend closely, although the mean stiffness stays relatively constant beyond four voids, instead of showing a slight downward trend towards nine voids.
Nevertheless, both models give stiffness predictions that closely follow the FE results.
Moreover, the decrease in variance for increasing microstructure size that is observed in the FEM predictions is reproduced also for unseen microstructure sizes.
With this, we show that there is a positive inductive bias in the microstructure geometry that our model benefits from.

\subsection{Computational cost}\label{sec:comp_cost}
Making a fair time comparison between using the GNN surrogate and a full \fetwo{} simulation is challenging for several reasons.
The GNN surrogate is implemented in Python using PyTorch, created for GPU execution, while our in-house FE solver is written in C++ and executed on a CPU.
We therefore focus on the relative scaling of both methods with the size of the microscale mesh, rather than on the absolute differences.
We compute all steps sequentially for both the FE and GNN approaches, without using parallelization or batching techniques.

\begin{figure}[htbp]
\centering
\includegraphics[width=0.5\textwidth]{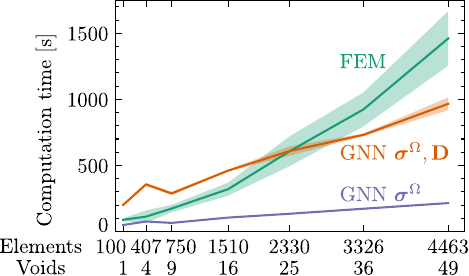}
\caption{Scaling of the computational cost between the GNN-based surrogate and FEM for an increasing microstructure size. We plot the mean and 95\% confidence interval over 5 runs.}
\label{fig:time_comparison}
\end{figure}

We perform multiscale simulations by applying tension to a macroscopic dogbone structure with holes.
We vary the microstructure size between runs to study the scaling of the computation time.
For each simulation, the strain paths and computational cost of solving the microscale model for all time steps are recorded, considering only the final Newton-Raphson steps that lead to macroscopic convergence.
We subject the surrogate model (base model A) to the same strain paths and record the resulting execution time.
The resulting computation times corresponding to the microstructure size are shown in Figure~\ref{fig:time_comparison}.
The results match the expectation that the GNN surrogate scales linearly with respect to the microstructure size, whereas FEM scales exponentially.
We also observe that computing the tangent stiffness using automatic differentiation is costly and scales worse compared to when only computing $\boldsymbol{\sigma}^{\Omega}$.
Therefore, other methods to compute the tangent stiffness, such as finite differences, could be considered instead.
Some hyperparameters of the GNN model, such as the number of MPLs, will also influence the computation time.
However, these are not expected to change how the GNN scales favorably with the microstructure size compared to full \fetwo{} simulations.

\section{Conclusion}\label{sec:conclusion}
State-of-the-art surrogate models for multiscale simulations directly replace the complete microscale simulation with a machine-learning model.
By doing so, the microscale physics and geometry are lost, and a large number of simulations from RVEs are required for training.
We presented a GNN-based surrogate that replaces only the solving of the microscopic boundary-value problem while keeping the microscopic material models and computational homogenization intact.
By obtaining all full-field quantities of interest the surrogate could be used interchangeably with an FE solver for any time step in a multiscale model.

We demonstrated the ability of our model to accurately predict microscopic strain and stress fields for monotonically increasing strain paths with nonlinearity caused by considering elasto-plastic materials.
We found that embedding the material model in the GNN is insufficient to capture complex load scenarios such as unloading while only being trained on monotonic loading.
Instead, we trained our model on a dataset with non-monotonic load paths, which improves its ability to predict unloading scenarios.
Embedding the material model inside the surrogate architecture leads to a significant performance increase.
While the stress directly depends on the strain, we found that including the stress in the loss function (in addition to the strain) not only leads to a significant improvement in the stress predictions but also generally improves the strain predictions.
The model is autoregressive and therefore potentially sensitive to instability, but we show that the error does not increase excessively when extrapolating beyond the number of time steps seen during training.
By training on arbitrary microstructures and predicting the stiffness of larger microstructures than when training the model, we show the potential for GNNs to extrapolate to larger microstructures.
This is possible by the inductive bias of the GNN, which learns to use the geometry of the microstructure to make predictions.
The proposed surrogate has potential to be faster than an FE analysis by scaling better with the number of elements in the microstructure.
We also expect these gains to be higher once our GNN code is further optimized and better integrated within existing FE code.

By keeping the multiscale nature of the problem while being faster than full \fetwo{} simulations, this model shows promise in accelerating simulations of multiscale mechanical models.
This is the first surrogate to integrate an elasto-plastic material into a GNN while retaining all microstructure quantities for a multiscale simulation.
With countless developments on GNNs, we expect that they will play an increasingly important role in the field of multiscale simulations, as they are naturally suited to model mesh-based systems.
We further expect that purely data-driven models will not be optimal, and that hybrid data-driven and physics-based models such as the one we show here will provide more powerful surrogates for multiscale simulations.

\section*{Acknowledgements}\label{sec:acknowledge}
This work is supported by the TU Delft AI Labs programme.
FM also acknowledges financial support from the Netherlands Organization for Scientific Research (NWO) under Vidi grant nr. 16464.


%
\bibliographystyle{elsarticle-num}
\bibliography{reflibproposal}

\begin{thebibliography}{10}
\expandafter\ifx\csname url\endcsname\relax
  \def\url#1{\texttt{#1}}\fi
\expandafter\ifx\csname urlprefix\endcsname\relax\def\urlprefix{URL }\fi
\expandafter\ifx\csname href\endcsname\relax
  \def\href#1#2{#2} \def\path#1{#1}\fi

\bibitem{ghavamian2019accelerating}
F.~Ghavamian, A.~Simone, Accelerating multiscale finite element simulations of
  history-dependent materials using a recurrent neural network, Computer
  Methods in Applied Mechanics and Engineering 357 (2019) 112594.

\bibitem{borkowski2022recurrent}
L.~Borkowski, C.~Sorini, A.~Chattopadhyay, Recurrent neural network-based
  multiaxial plasticity model with regularization for physics-informed
  constraints, Computers \& Structures 258 (2022) 106678.

\bibitem{wu2020recurrent}
L.~Wu, N.~G. Kilingar, L.~Noels, et~al., A recurrent neural network-accelerated
  multi-scale model for elasto-plastic heterogeneous materials subjected to
  random cyclic and non-proportional loading paths, Computer Methods in Applied
  Mechanics and Engineering 369 (2020) 113234.

\bibitem{wang2022deep}
J.-J. Wang, C.~Wang, J.-S. Fan, Y.~Mo, A deep learning framework for
  constitutive modeling based on temporal convolutional network, Journal of
  Computational Physics 449 (2022) 110784.

\bibitem{abueidda2021deep}
D.~W. Abueidda, S.~Koric, N.~A. Sobh, H.~Sehitoglu, Deep learning for
  plasticity and thermo-viscoplasticity, International Journal of Plasticity
  136 (2021) 102852.

\bibitem{wang2020general}
C.~Wang, L.-y. Xu, J.-s. Fan, A general deep learning framework for
  history-dependent response prediction based on ua-seq2seq model, Computer
  Methods in Applied Mechanics and Engineering 372 (2020) 113357.

\bibitem{linka2021constitutive}
K.~Linka, M.~Hillg{\"a}rtner, K.~P. Abdolazizi, R.~C. Aydin, M.~Itskov, C.~J.
  Cyron, Constitutive artificial neural networks: a fast and general approach
  to predictive data-driven constitutive modeling by deep learning, Journal of
  Computational Physics 429 (2021) 110010.

\bibitem{klein2022finite}
D.~K. Klein, R.~Ortigosa, J.~Mart{\'\i}nez-Frutos, O.~Weeger, Finite
  electro-elasticity with physics-augmented neural networks, Computer Methods
  in Applied Mechanics and Engineering 400 (2022) 115501.

\bibitem{masi2021thermodynamics}
F.~Masi, I.~Stefanou, P.~Vannucci, V.~Maffi-Berthier, Thermodynamics-based
  artificial neural networks for constitutive modeling, Journal of the
  Mechanics and Physics of Solids 147 (2021) 104277.

\bibitem{maia2022physically}
M.~Maia, I.~Rocha, P.~Kerfriden, F.~van~der Meer, Physically recurrent neural
  networks for path-dependent heterogeneous materials: embedding constitutive
  models in a data-driven surrogate, arXiv preprint arXiv:2209.07320 (2022).

\bibitem{liu2019deep}
Z.~Liu, C.~Wu, M.~Koishi, A deep material network for multiscale topology
  learning and accelerated nonlinear modeling of heterogeneous materials,
  Computer Methods in Applied Mechanics and Engineering 345 (2019) 1138--1168.

\bibitem{rocha2023machine}
I.~Rocha, P.~Kerfriden, F.~van~der Meer, Machine learning of evolving
  physics-based material models for multiscale solid mechanics, arXiv preprint
  arXiv:2301.13547 (2023).

\bibitem{raissi2019physics}
M.~Raissi, P.~Perdikaris, G.~E. Karniadakis, Physics-informed neural networks:
  A deep learning framework for solving forward and inverse problems involving
  nonlinear partial differential equations, Journal of Computational physics
  378 (2019) 686--707.

\bibitem{karniadakis2021physics}
G.~E. Karniadakis, I.~G. Kevrekidis, L.~Lu, P.~Perdikaris, S.~Wang, L.~Yang,
  Physics-informed machine learning, Nature Reviews Physics 3~(6) (2021)
  422--440.

\bibitem{PIGalerkingGNN}
H.~Gao, M.~J. Zahr, J.-X. Wang, Physics-informed graph neural galerkin
  networks: A unified framework for solving pde-governed forward and inverse
  problems, Computer Methods in Applied Mechanics and Engineering 390 (2022)
  114502.

\bibitem{lu2019deeponet}
L.~Lu, P.~Jin, G.~E. Karniadakis, Deeponet: Learning nonlinear operators for
  identifying differential equations based on the universal approximation
  theorem of operators, arXiv preprint arXiv:1910.03193 (2019).

\bibitem{li2020fourier}
Z.~Li, N.~Kovachki, K.~Azizzadenesheli, B.~Liu, K.~Bhattacharya, A.~Stuart,
  A.~Anandkumar, Fourier neural operator for parametric partial differential
  equations, arXiv preprint arXiv:2010.08895 (2020).

\bibitem{li2020neural}
Z.~Li, N.~Kovachki, K.~Azizzadenesheli, B.~Liu, K.~Bhattacharya, A.~Stuart,
  A.~Anandkumar, Neural operator: Graph kernel network for partial differential
  equations, arXiv preprint arXiv:2003.03485 (2020).

\bibitem{cang2018improving}
R.~Cang, H.~Li, H.~Yao, Y.~Jiao, Y.~Ren, Improving direct physical properties
  prediction of heterogeneous materials from imaging data via convolutional
  neural network and a morphology-aware generative model, Computational
  Materials Science 150 (2018) 212--221.

\bibitem{abueidda2019prediction}
D.~W. Abueidda, M.~Almasri, R.~Ammourah, U.~Ravaioli, I.~M. Jasiuk, N.~A. Sobh,
  Prediction and optimization of mechanical properties of composites using
  convolutional neural networks, Composite Structures 227 (2019) 111264.

\bibitem{krokos2022bayesian}
V.~Krokos, V.~Bui~Xuan, S.~Bordas, P.~Young, P.~Kerfriden, A bayesian
  multiscale cnn framework to predict local stress fields in structures with
  microscale features, Computational Mechanics 69~(3) (2022) 733--766.

\bibitem{gupta2023accelerated}
A.~Gupta, A.~Bhaduri, L.~Graham-Brady, Accelerated multiscale mechanics
  modeling in a deep learning framework, Mechanics of Materials (2023) 104709.

\bibitem{aldakheel2023efficient}
F.~Aldakheel, E.~S. Elsayed, T.~I. Zohdi, P.~Wriggers, Efficient multiscale
  modeling of heterogeneous materials using deep neural networks, Computational
  Mechanics (2023) 1--17.

\bibitem{vijayaraghavan2023data}
S.~Vijayaraghavan, L.~Wu, L.~Noels, S.~Bordas, S.~Natarajan, L.~A. Beex, A
  data-driven reduced-order surrogate model for entire elastoplastic
  simulations applied to representative volume elements, Scientific Reports
  13~(1) (2023) 12781.

\bibitem{krokos2022graph}
V.~Krokos, S.~Bordas, P.~Kerfriden, A graph-based probabilistic geometric deep
  learning framework with online physics-based corrections to predict the
  criticality of defects in porous materials, arXiv preprint arXiv:2205.06562
  (2022).

\bibitem{pfaff2020learning}
T.~Pfaff, M.~Fortunato, A.~Sanchez-Gonzalez, P.~W. Battaglia, Learning
  mesh-based simulation with graph networks, arXiv preprint arXiv:2010.03409
  (2020).

\bibitem{li2018deeper}
Q.~Li, Z.~Han, X.-M. Wu, Deeper insights into graph convolutional networks for
  semi-supervised learning, in: Proceedings of the AAAI conference on
  artificial intelligence, Vol.~32, 2018.

\bibitem{li2020multipole}
Z.~Li, N.~Kovachki, K.~Azizzadenesheli, B.~Liu, A.~Stuart, K.~Bhattacharya,
  A.~Anandkumar, Multipole graph neural operator for parametric partial
  differential equations, Advances in Neural Information Processing Systems 33
  (2020) 6755--6766.

\bibitem{multiscaleGNN}
M.~Lino, C.~Cantwell, A.~A. Bharath, S.~Fotiadis, Simulating continuum
  mechanics with multi-scale graph neural networks, arXiv preprint
  arXiv:2106.04900 (2021).

\bibitem{cao2022bi}
Y.~Cao, M.~Chai, M.~Li, C.~Jiang, Bi-stride multi-scale graph neural network
  for mesh-based physical simulation, arXiv preprint arXiv:2210.02573 (2022).

\bibitem{fortunato2022multiscale}
M.~Fortunato, T.~Pfaff, P.~Wirnsberger, A.~Pritzel, P.~Battaglia, Multiscale
  meshgraphnets, arXiv preprint arXiv:2210.00612 (2022).

\bibitem{gladstone2023gnn}
R.~J. Gladstone, H.~Rahmani, V.~Suryakumar, H.~Meidani, M.~D'Elia, A.~Zareei,
  Gnn-based physics solver for time-independent pdes, arXiv preprint
  arXiv:2303.15681 (2023).

\bibitem{brandstetter2022message}
J.~Brandstetter, D.~Worrall, M.~Welling, Message passing neural pde solvers,
  arXiv preprint arXiv:2202.03376 (2022).

\bibitem{maurizi2022predicting}
M.~Maurizi, C.~Gao, F.~Berto, Predicting stress, strain and deformation fields
  in materials and structures with graph neural networks, Scientific Reports
  12~(1) (2022) 21834.

\bibitem{vlassis2023geometric}
N.~N. Vlassis, W.~Sun, Geometric learning for computational mechanics part ii:
  Graph embedding for interpretable multiscale plasticity, Computer Methods in
  Applied Mechanics and Engineering 404 (2023) 115768.

\bibitem{frankel2022mesh}
A.~L. Frankel, C.~Safta, C.~Alleman, R.~E. Jones, Mesh-based graph
  convolutional neural networks for modeling materials with microstructure,
  Journal of Machine Learning for Modeling and Computing 3~(1) (2022).

\bibitem{hamilton2020graph}
W.~L. Hamilton, Graph representation learning, Synthesis Lectures on Artifical
  Intelligence and Machine Learning 14~(3) (2020) 1--159.

\bibitem{battaglia2018relational}
P.~W. Battaglia, J.~B. Hamrick, V.~Bapst, A.~Sanchez-Gonzalez, V.~Zambaldi,
  M.~Malinowski, A.~Tacchetti, D.~Raposo, A.~Santoro, R.~Faulkner, et~al.,
  Relational inductive biases, deep learning, and graph networks, arXiv
  preprint arXiv:1806.01261 (2018).

\bibitem{vlassis2020geometric}
N.~N. Vlassis, R.~Ma, W.~Sun, Geometric deep learning for computational
  mechanics part i: Anisotropic hyperelasticity, Computer Methods in Applied
  Mechanics and Engineering 371 (2020) 113299.

\bibitem{nguyen2020jive}
C.~Nguyen-Thanh, V.~P. Nguyen, A.~de~Vaucorbeil, T.~K. Mandal, J.-Y. Wu, Jive:
  An open source, research-oriented c++ library for solving partial
  differential equations, Advances in Engineering Software 150 (2020) 102925.

\bibitem{gmsh}
C.~Geuzaine, J.-F. Remacle, Gmsh: A 3-d finite element mesh generator with
  built-in pre-and post-processing facilities, International journal for
  numerical methods in engineering 79~(11) (2009) 1309--1331.

\bibitem{Fey/Lenssen/2019}
M.~Fey, J.~E. Lenssen, Fast graph representation learning with {PyTorch
  Geometric}, in: ICLR Workshop on Representation Learning on Graphs and
  Manifolds, 2019.

\bibitem{DHPC2022}
{D}elft {H}igh {P}erformance {C}omputing~{C}entre ({DHPC}), {D}elft{B}lue
  {S}upercomputer ({P}hase 1),
  \url{https://www.tudelft.nl/dhpc/ark:/44463/DelftBluePhase1} (2022).

\bibitem{mozaffar2019deep}
M.~Mozaffar, R.~Bostanabad, W.~Chen, K.~Ehmann, J.~Cao, M.~Bessa, Deep learning
  predicts path-dependent plasticity, Proceedings of the National Academy of
  Sciences 116~(52) (2019) 26414--26420.

\bibitem{logarzo2021smart}
H.~J. Logarzo, G.~Capuano, J.~J. Rimoli, Smart constitutive laws: Inelastic
  homogenization through machine learning, Computer methods in applied
  mechanics and engineering 373 (2021) 113482.

\end{thebibliography}

\appendix

\end{document}